\documentclass[journal,10pt,twoside]{IEEEtran}
\usepackage{comment}
\usepackage{graphicx}
\usepackage{float}
\usepackage{subcaption}  
\usepackage{amssymb} 

\usepackage{hyperref}
\usepackage{url}
\usepackage{booktabs}
\usepackage[table]{xcolor}
\usepackage{pifont}
\usepackage{hyperref}
\usepackage{titlesec}

\titleclass{\subsubsubsection}{straight}[\subsubsection]

\newcounter{subsubsubsection}[subsubsection]
\renewcommand\thesubsubsubsection{\thesubsubsection.\arabic{subsubsubsection}}

\titleformat{\subsubsubsection}[hang]
{\normalfont\normalsize}             
{\thesubsubsubsection}               
{1em}                                
{}

\titlespacing*{\subsubsubsection}
{0pt}{3.25ex plus 1ex minus .2ex}{1.5ex plus .2ex}

\title{SFlexRCA: Lightweight, Scalable, and Flexible Root Cause Analysis for IIoT Edge Systems}

\begin{document}

\author{
	Amr M. Zaki,
	Farhoud Jafari,
	Honggeun Ji,
	Komal Sarda,
	Marin Litoiu%
	\thanks{
		Amr M. Zaki (corresponding author), Farhoud Jafari, Honggeun Ji, Komal Sarda, and Marin Litoiu are with the Lassonde School of Engineering, York University, Toronto, ON, Canada.
		(e-mail: amrzaki@yorku.ca; farhoud@yorku.ca; gihong96@yorku.ca; komal253@yorku.ca; mlitoiu@yorku.ca).
	}
}
\maketitle

\begin{abstract}
	Industrial Internet of Things (IIoT) systems generate high-dimensional sensor telemetry from interconnected components, where faults can propagate across the system. Existing root cause analysis (RCA) approaches often depend on incomplete dependency or observability information, while non-causal methods may require abundant normal-operation data and causal approaches can incur substantial computational and parameter costs.
	To address these challenges, we propose SFlexRCA (Scalable and Flexible Root Cause Analysis), a topology-free RCA framework designed for resource-constrained IIoT environments. SFlexRCA transforms multivariate telemetry into compact orthogonal representations and applies shared lightweight linear modeling, avoiding explicit graph construction, message passing, and per-variable or lag-specific parameter growth.
	We evaluate SFlexRCA on three publicly available IIoT datasets, BATADAL, SWaT, and WADI, spanning different numbers of monitored variables, temporal characteristics, and training-data regimes. SFlexRCA is compared with 10 statistical, causal, and non-causal baselines in terms of RCA accuracy, training efficiency, inference latency, and memory consumption. In addition, inference efficiency and memory consumption are evaluated on Raspberry Pi 3 and Raspberry Pi 5, while energy consumption is additionally measured on Raspberry Pi 5.
	We further investigate temporal-context sensitivity, architectural and loss components, and alternative representations. SFlexRCA achieves the best AC@1 on BATADAL (0.476) and the best AC@5 on SWaT (0.367), while using only 279.4K and 108.1K parameters, respectively. Notably, SFlexRCA maintains strong localization performance on BATADAL despite its limited normal-operation training data, while its compact shared architecture avoids the parameter growth associated with causal and graph-based approaches. On Raspberry Pi 5, SFlexRCA achieves 20.3, 8.2, and 9.3~ms/sample on BATADAL, SWaT, and WADI, respectively.
	These results demonstrate that SFlexRCA provides a practical alternative to computationally intensive causal and graph-based RCA approaches, achieving strong localization with substantially smaller models while remaining effective under limited training data. Its lightweight shared architecture further enables efficient deployment on resource-constrained IIoT edge devices. The SFlexRCA code is available at \href{https://github.com/theamrzaki/RootCause-Analysis-Correlation-Attentive-Modeling}{https://github.com/theamrzaki/RootCause-Analysis-Correlation-Attentive-Modeling}.
\end{abstract}


\section{Introduction}

As industrial processes increasingly integrate networked sensors, actuators, controllers, and computing platforms, Industrial Internet of Things (IIoT) systems have become critical to modern water treatment, energy, manufacturing, and other cyber-physical infrastructures~\cite{iotcadis361}. These systems continuously generate heterogeneous time-series measurements, making data-driven monitoring essential for operational reliability and safety~\cite{fadsd-348,TSAD-347}. However, failures can propagate across physically and logically coupled components, producing correlated abnormal behavior across multiple sensors and subsystems~\cite{causality-neurips-297}. Identifying the source of such abnormal behavior is therefore essential for reliable and resilient IIoT operation~\cite{iot-2-of-361}.

Anomaly detection can identify deviations from normal behavior~\cite{iot-1-of-361}, but does not determine which component or process variable initiated the failure~\cite{RCAEval_302}. Root cause analysis (RCA) addresses this limitation by identifying the most likely sources of abnormal behavior and supporting corrective actions~\cite{CUTS_PLUS}. This task is particularly challenging in industrial environments because sensors, actuators, control loops, and physical subsystems are strongly coupled. A fault originating from one component can propagate through the process and appear as correlated anomalies across many measurements, making it difficult to distinguish the initiating fault from downstream effects.
Existing RCA approaches face several practical limitations in IIoT settings. Multimodal methods may rely on logs and distributed traces that are incomplete or unavailable in industrial environments~\cite{283-43-Lee2023Eadro,Art-319,270-MSTGAD,271-withLLM,263-complexJournal}. Recent IoT and industrial approaches have increasingly explored causal structure discovery, causal interpretability, and dynamic causal graphs to capture dependencies among system variables~\cite{Cao2026Causal,Zeng2025Causal,Tang2025LLMDCGRCA}. Other approaches combine reinforcement learning and graph-based causal diagnosis for root cause localization~\cite{PA-Rank2025}. Some methods further incorporate large language models (LLMs) to assist causal reasoning and root cause localization~\cite{Tang2025LLMDCGRCA}. While these approaches can exploit rich dependency information, explicit causal structures, graph construction, reinforcement-learning-based discovery, and LLM-based components can introduce additional computational and memory requirements, making deployment challenging on resource-constrained edge devices. Moreover, graph message passing~\cite{CUTS_PLUS}, independent models for individual variables~\cite{cLSTM}, and lag- or context-dependent architectures~\cite{GVAR,285-encoder-decoder} can further increase parameter and computational costs. These constraints are particularly important for distributed IoT systems, where efficient computation, energy use, and resource allocation remain central challenges for edge intelligence~\cite{Zaki2025QBATO,GreeningEdgeAI2025,FLARE2026}. At the same time, non-causal forecasting approaches can provide compact representations but may require substantial normal-operation data for reliable localization, which can be difficult to obtain in industrial settings.
These limitations expose a gap between compact non-causal models and increasingly sophisticated but computationally demanding dependency-aware approaches. This motivates the question: Can root cause localization retain the benefits of dependency-aware modeling without requiring explicit dependency graphs, computationally intensive causal-discovery mechanisms, LLM-based components, or expensive per-variable and graph-based architectures, while remaining effective across different data regimes and practical for edge deployment?

We address this gap with \textbf{SFlexRCA (Scalable and Flexible Root Cause Analysis)}, a lightweight and topology-free framework for industrial root cause localization. SFlexRCA represents multivariate telemetry in a compact orthogonal space and uses shared linear modeling to capture temporal and cross-variable patterns. Industrial measurements often contain redundancy, correlation, and shared temporal structure, making compact representations a promising alternative to explicitly modeling every variable and dependency~\cite{Olinear_336,vlinear_337,276-Fits,FreTS}. By operating in a shared representation space, SFlexRCA avoids explicit dependency graphs, message passing, and independent per-variable models, thereby limiting parameter growth as telemetry dimensionality and temporal context increase. Its modular architecture also allows the model to adapt to different industrial systems.

We evaluate SFlexRCA on \textbf{SWaT, WADI, and BATADAL}~\cite{295-mathur2016-swat,wadi-34-of-301,BATADAL}, covering different telemetry dimensionalities and training-data regimes. We assess RCA accuracy, parameter complexity, training and inference efficiency, and memory consumption on both \textbf{Raspberry Pi 3 and Raspberry Pi 5}, while evaluating energy consumption on Raspberry Pi 5. SFlexRCA achieves the best AC@1 on BATADAL and the best AC@5 on SWaT while using substantially fewer trainable parameters than competing approaches, demonstrating its practical scalability without the parameter growth associated with dependency-aware causal and graph-based models. It also remains effective under limited normal-operation training data, as observed on BATADAL. We further examine temporal-context sensitivity, architectural and loss components, alternative representation methods, and the learned representations underlying localization.

The main contributions of this work are:
\begin{itemize}
	\item A lightweight, topology-free RCA framework based on orthogonal representations and shared linear modeling, avoiding explicit dependency graphs and per-variable or per-window architectures.
	\item An adaptive architecture that limits parameter growth with telemetry dimensionality and temporal context while supporting different industrial data regimes.
	\item A comprehensive evaluation across SWaT, WADI, and BATADAL, including accuracy, computational and memory efficiency, and edge energy consumption.
	\item Ablation and sensitivity analyses examining the proposed components, temporal context, representation choices, and localization behavior.
\end{itemize}

This work is guided by the following research questions:

\textbf{RQ1:} How does SFlexRCA balance root cause localization accuracy and model complexity in terms of trainable parameters?

\textbf{RQ2:} How does SFlexRCA compare in training efficiency, memory consumption, and inference latency on Raspberry Pi 3 and 5?

\textbf{RQ3:} What are the contributions of SFlexRCA's architectural and loss components to its localization performance?

\textbf{RQ4:} How does temporal context affect SFlexRCA's root cause localization accuracy?

\textbf{RQ5:} How does SFlexRCA's orthogonal representation compare with alternative representations in terms of localization accuracy and latent-space separability?

This paper is organized as follows. Section 2 reviews related work. Section 3 presents the problem formulation and SFlexRCA. Section 4 presents the experimental evaluation. Section 5 includes threat analysis and discussion. Section 6 concludes the paper.

\section{Related Work}
RCA in multivariate systems lies at the intersection of performance engineering and system diagnosis, where the goal extends beyond anomaly detection toward scalable and operationally practical fault localization.

\begin{table}[t]

    \scriptsize
    \centering
    \renewcommand{\arraystretch}{1.15}
    \setlength{\tabcolsep}{4pt}
    \caption{Comparison of Representative Root Cause Analysis and Efficient Time-Series Modeling Approaches.}
    \label{tab:related_works_rca}

    \begin{tabular}{|p{2.0cm}|p{0.7cm}|p{2.5cm}|p{0.8cm}|p{1.3cm}|}
        \hline
        Method & Telemetry & Architecture & Edge Evaluation & Target \\
        \hline

        \multicolumn{5}{|l|}{\cellcolor{gray} Statistical RCA} \\
        \hline

        $\epsilon$-Diagnosis~\cite{286-Shan2019-EpsilonDiagnosis}
        & M
        & Two-sample testing
        & --
        & Microservices
        \\ \hline

        RCD~\cite{286-Ikram2022-RCD}
        & M
        & Conditional independence
        & --
        & Microservices
        \\ \hline

        BARO~\cite{288-BARO}
        & M
        & Robust statistical scoring
        & --
        & Microservices
        \\ \hline

        \multicolumn{5}{|l|}{\cellcolor{gray} Multimodal RCA} \\
        \hline

        ART~\cite{Art-319}
        & L+M+T
        & Transformer + GRU
        & --
        & Distributed systems
        \\ \hline

        Eadro~\cite{283-43-Lee2023Eadro}
        & L+M+T
        & GAT
        & --
        & Distributed systems
        \\ \hline

        MSTGAD~\cite{270-MSTGAD}
        & L+M+T
        & Spatial-temporal Transformer
        & --
        & Distributed systems
        \\ \hline

        LLM4MST~\cite{271-withLLM}
        & L+M+T
        & GPT-2
        & --
        & Distributed systems
        \\ \hline

        \multicolumn{5}{|l|}{\cellcolor{gray} Neural Dependency and Causal RCA} \\
        \hline

        cMLP/cLSTM~\cite{cLSTM}
        & M
        & Per-variable MLP/LSTM
        & --
        & Causal RCA
        \\ \hline

        CUTS~\cite{cuts_one}
        & M
        & GNN + imputation
        & --
        & Causal RCA
        \\ \hline

        GVAR~\cite{GVAR}
        & M
        & Per-window MLP
        & --
        & Causal RCA
        \\ \hline

        AERCA~\cite{285-encoder-decoder}
        & M
        & Per-window MLP
        & --
        & Causal RCA
        \\ \hline

        CUTS+~\cite{CUTS_PLUS}
        & M
        & Message passing
        & --
        & Causal RCA
        \\ \hline

        Cao et al.~\cite{Cao2026Causal}
        & M
        & Causal structure discovery
        & --
        & Industrial monitoring
        \\ \hline

        Tang et al.~\cite{Tang2025LLMDCGRCA}
        & M
        & Dynamic causal graph + LLM
        & --
        & IoT RCA
        \\ \hline
        \multicolumn{5}{|l|}{\cellcolor{gray} Non-Causal Forecasting} \\
        \hline

        iTransformer~\cite{itransformer}
        & M
        & Inverted Transformer
        & --
        & Forecasting
        \\ \hline

        TimeMixer++~\cite{timemixer++}
        & M
        & Multiscale mixing
        & --
        & Forecasting
        \\ \hline

        \multicolumn{5}{|l|}{\cellcolor{gray} Efficient Linear and Edge Time-Series Methods} \\
        \hline

        FITS~\cite{276-Fits}
        & M
        & Linear / frequency
        & --
        & Forecasting
        \\ \hline

        FilterNet~\cite{277-filternet}
        & M
        & Linear / frequency
        & --
        & Forecasting
        \\ \hline

        OLinear~\cite{olinear}
        & M
        & Orthogonal linear
        & --
        & Forecasting
        \\ \hline

        VLinear~\cite{vlinear_337}
        & M
        & Linear projection
        & --
        & Forecasting
        \\ \hline

        FADSD~\cite{fadsd-348}
        & M
        & Lightweight statistical
        & RPi / STM32
        & Anomaly detection
        \\ \hline

        LFTSAD~\cite{TSAD-347}
        & M
        & Lightweight MLP
        & RPi / Jetson
        & Anomaly detection
        \\ \hline

        \rowcolor{green!20}
        \textbf{SFlexRCA (Ours)}
        & \textbf{M}
        & \textbf{Orthogonal projection + shared linear modeling}
        & \textbf{RPi 3 / RPi 5}
        & \textbf{IIoT RCA}
        \\ \hline

    \end{tabular}

    \begin{flushleft}
        \tiny
        M: Metrics/sensor measurements, L: Logs, T: Traces.
    \end{flushleft}

\end{table}

\subsection{Root Cause Analysis}

Existing RCA approaches can be broadly categorized into statistical, multimodal, non-causal forecasting, and neural dependency-based methods, each making different assumptions about the available observability, training data, and system structure.


Statistical RCA methods primarily identify abnormal dependencies or distributional shifts among monitoring signals. $\epsilon$-Diagnosis~\cite{286-Shan2019-EpsilonDiagnosis} uses two-sample statistical testing, while RCD~\cite{286-Ikram2022-RCD} employs conditional independence tests to infer causal relationships among system metrics. BARO~\cite{288-BARO} further improves robustness through median- and IQR-based scoring. These methods are computationally efficient and interpretable, but are primarily designed for microservice and software-system telemetry and can struggle to capture the complex temporal dependencies and nonlinear interactions characteristic of industrial processes.


A second line of work exploits multimodal observability by jointly analyzing metrics, logs, and traces. Approaches such as ART, Eadro, MSTGAD, and LLM4MST~\cite{283-43-Lee2023Eadro,Art-319,270-MSTGAD,271-withLLM,263-complexJournal} leverage complementary information across modalities to improve fault localization. However, these approaches assume the availability of sufficiently complete and aligned multimodal telemetry, which may be impractical in industrial environments dominated by sensor and control measurements. Industrial systems may lack distributed traces or consistent logs across heterogeneous controllers and devices, making single-modality approaches more suitable for partially observable and resource-constrained deployments.


Non-causal forecasting approaches provide an alternative by learning temporal patterns directly from multivariate observations without explicitly modeling causal or dependency structures. Recent models such as TimeMixer++~\cite{timemixer++} and iTransformer~\cite{itransformer} can capture complex temporal and cross-variable patterns and have demonstrated strong forecasting capability. However, their effectiveness for root cause localization can depend on the availability of abundant normal-operation data. This limitation is particularly relevant to IIoT environments, where collecting representative normal-operation data can be difficult due to operational costs, system changes, or limited periods of stable operation. Moreover, forecasting accuracy does not necessarily translate directly into reliable root cause localization, since accurate prediction of system behavior does not explicitly identify the variable responsible for an observed failure. Thus, while non-causal models can provide attractive computational efficiency, their data requirements and indirect localization mechanism can limit their applicability to data-limited diagnostic scenarios and resource-constrained deployments.


Neural dependency-based methods instead attempt to learn temporal or causal relationships directly from multivariate observations. Methods such as cMLP and cLSTM~\cite{cLSTM} use neural networks associated with individual variables to recover dependency structures. CUTS~\cite{cuts_one} addresses irregular and incomplete time-series data through a neural Granger-causality framework that jointly imputes missing observations and learns causal graphs using graph neural networks. Recent work has further investigated causal structure discovery for industrial processes, including nonlinear systems with complex noise~\cite{Cao2026Causal}, while IoT-oriented RCA approaches have employed dynamic causal graphs to model evolving dependencies and have incorporated LLMs to assist causal reasoning and root cause localization~\cite{Tang2025LLMDCGRCA}. GVAR~\cite{GVAR} and AERCA~\cite{285-encoder-decoder} use deep temporal architectures that process historical monitoring windows, requiring deep models to be repeatedly applied across temporal windows during training and inference. CUTS+~\cite{CUTS_PLUS} further employs message passing over learned representations to model interactions among variables. These approaches provide more direct dependency-aware mechanisms for localization and can remain effective when training data are limited. However, learning explicit causal structures, using per-variable neural networks, repeatedly processing temporal windows, or performing graph-based message passing can introduce substantial computational and memory overhead. LLM-assisted causal approaches further increase these resource requirements, making deployment challenging on resource-constrained IIoT edge devices, particularly as the number of monitored variables and temporal context increase.


Overall, existing RCA approaches exhibit a fundamental trade-off between data requirements, dependency modeling, and computational cost. Non-causal forecasting methods can provide efficient temporal modeling but may require substantial normal-operation data and do not directly model root-cause dependencies. Causal and neural dependency-based methods provide more explicit dependency-aware localization but often rely on computationally intensive per-variable, temporal, graph-based, or LLM-assisted architectures. In contrast, SFlexRCA targets topology-free RCA from a single modality of industrial telemetry by combining orthogonal representations with shared lightweight linear components. This design captures temporal and cross-variable information without explicitly constructing dependency graphs, using message passing or LLMs, or relying on per-variable and lag-specific architectures, enabling efficient root cause localization across different training-data regimes and supporting deployment on resource-constrained IIoT edge devices.

\subsection{Orthogonal and Linear Temporal Modeling}

Recent advances in high-dimensional time-series modeling suggest that structured transformations can simplify temporal learning while reducing computational cost. Frequency-domain methods such as FITS~\cite{276-Fits}, FEDformer~\cite{fedformer}, FilterNet~\cite{277-filternet}, and FourierGNN~\cite{278-FourierGNN} demonstrate the effectiveness of alternative representations for modeling long-range temporal dependencies. More recently, OLinear~\cite{olinear} and VLinear~\cite{vlinear_337} show that dataset-specific linear projections can achieve competitive performance with substantially lower computational cost, suggesting that complex temporal structure can be captured in transformed spaces without deep nonlinear architectures.
Recent efforts have also investigated efficient anomaly detection directly on resource-constrained devices. For example, \cite{fadsd-348,TSAD-347} evaluate lightweight time-series anomaly detection models on edge platforms, including Raspberry Pi, STM32, and NVIDIA Jetson Xavier. While these studies demonstrate the feasibility of deploying anomaly detection at the edge, they focus on identifying anomalous behavior rather than determining its underlying root cause.

In contrast, SFlexRCA applies lightweight orthogonal projections and shared linear components to root cause localization. By avoiding explicit dependency graphs, message passing, and per-variable or per-window architectures, SFlexRCA enables topology-free RCA from single-modality industrial telemetry while targeting efficient deployment on resource-constrained IIoT edge devices.

\section{SFlexRCA}
\subsection{Problem Formulation}

Let
$X \in \mathbb{R}^{T \times P}$
denote a multivariate IIoT telemetry window containing $P$ sensor variables observed over $T$ time steps. Given normal operating data, SFlexRCA learns the temporal and cross-variable behavior of the system and uses deviations from this learned behavior to localize potential root causes.

Given an input window $X$, SFlexRCA produces a future-state prediction $\hat{Y}$ and an implicit variable-interaction representation $C$:

\begin{equation}
    \hat{Y}, C = \mathrm{SFlexRCA}(X),
\end{equation}

where $\hat{Y}$ represents the predicted future system state and
$C \in \mathbb{R}^{T \times P \times P}$
captures pairwise interactions inferred from the learned latent representations. Unlike topology-aware RCA approaches, SFlexRCA does not require a predefined dependency graph, process topology, or explicit communication structure among monitored variables.

\subsection{SFlexRCA Architecture}

\begin{figure*}[t]
	\centering
	\includegraphics[width=0.9\textwidth]{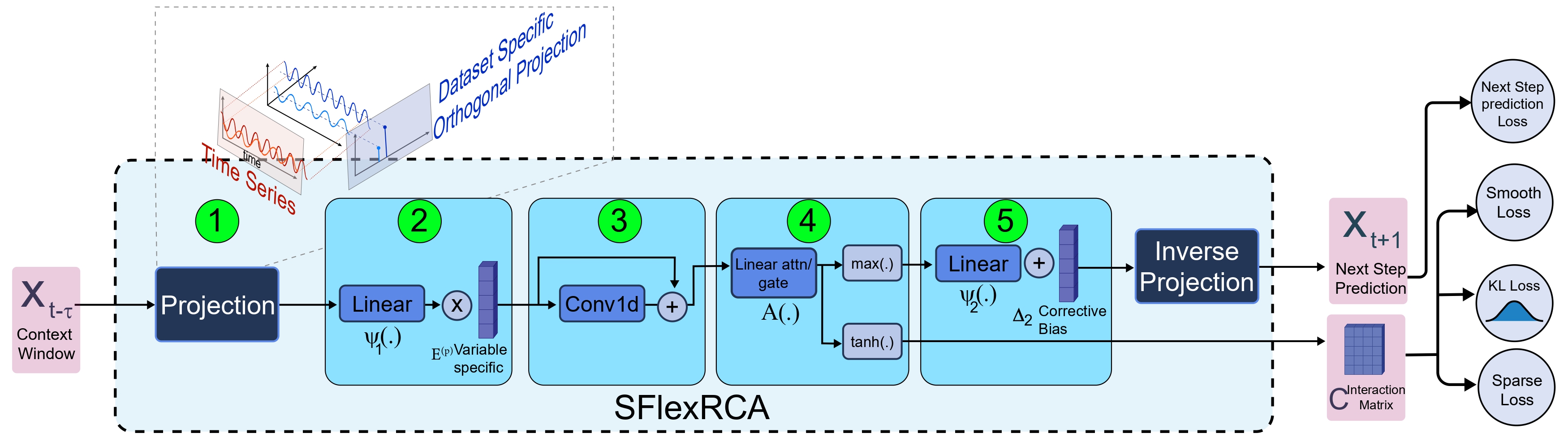}
	\caption{SFlexRCA framework for unsupervised root-cause analysis in IIoT systems.}
	\label{fig:sflexrca}
\end{figure*}

SFlexRCA is designed for single-modality IIoT telemetry, where large numbers of heterogeneous sensor measurements must be modeled efficiently under the computational constraints of edge devices. As illustrated in Fig.~\ref{fig:sflexrca}, SFlexRCA consists of five main stages: (i) orthogonal representation, (ii) shared variable encoding, (iii) temporal context modeling, (iv) context refinement and cross-variable interaction modeling, and (v) temporal aggregation and prediction. The architecture uses lightweight shared components rather than per-variable neural networks or graph-based message-passing architectures.

\textbf{(1) Orthogonal Representation.}

IIoT monitoring data typically contain highly correlated measurements and substantial temporal redundancy across sensor windows. Directly learning from the raw temporal coordinates can therefore require the downstream RCA model to repeatedly capture dependencies that are already present in the normal operating data. To provide a compact temporal representation, SFlexRCA constructs a dataset-specific orthogonal basis from normal IIoT telemetry following~\cite{olinear}.

For each monitored variable $p$, we compute its temporal correlation matrix over the normal training windows and obtain a shared correlation structure by averaging across variables:

\begin{equation}
	\bar{\Sigma}
	=
	\frac{1}{P}
	\sum_{p=1}^{P}
	\mathrm{Corr}(X^{(p)}),
\end{equation}

where $X^{(p)}$ contains the temporal observations of variable $p$. The eigenvectors of $\bar{\Sigma}$ define the orthogonal basis $Q$:

\begin{equation}
	\bar{\Sigma}
	=
	Q^\top\Lambda Q,
	\qquad
	Q^\top Q=I.
\end{equation}

The normalized telemetry is then projected as

\begin{equation}
	X_{\mathrm{orth}} = Q\tilde{X}.
\end{equation}

Because $Q$ is orthogonal, the transformation preserves the Euclidean geometry of the telemetry while expressing temporal variations along orthogonal directions. The basis is computed once using normal training data and remains fixed during model training, introducing no additional trainable parameters. This design is particularly suitable for IIoT settings because the representation separates recurring temporal structure from the subsequent learning of cross-variable interactions, allowing the latter to be modeled using a shared lightweight architecture rather than independent models for each sensor. The inverse transformation $Q^\top$ is used to recover predictions in the original measurement space.

\textbf{(2) Shared Variable Encoding.}

After transformation, each scalar observation is mapped into a shared latent space of dimension $D$ using the same projection function across all variables. To distinguish heterogeneous sensor variables, SFlexRCA employs a learnable variable embedding $E^{(p)}$ for each monitored variable $p$~\cite{vlinear_337}. In the default configuration, the latent representation is constructed as

\begin{equation}
	H_t^{(p)}
	=
	\psi(X_{\mathrm{orth},t}^{(p)})
	\odot E^{(p)}.
\end{equation}

This shared parameterization encodes variable-specific characteristics without requiring a separate neural network for each variable, keeping the projection parameters independent of the number of monitored variables~\cite{cLSTM,cuts_one}.

\textbf{(3) Temporal Context Modeling.}

To capture local temporal dependencies, SFlexRCA optionally applies a lightweight depthwise one-dimensional convolution:

\begin{equation}
	H
	\leftarrow
	H + \mathrm{Conv1D}(H),
\end{equation}

where the convolution operates independently over latent channels, providing local temporal mixing with minimal additional parameters. Its contribution is evaluated through architectural ablation.

\textbf{(4) Context Refinement and Cross-Variable Interaction Modeling.}

The temporally encoded representation is refined using a lightweight context mechanism:

\begin{equation}
	H' = \mathcal{A}(H),
\end{equation}

where $\mathcal{A}(\cdot)$ denotes either gated modulation or linear attention~\cite{linearattention}. The latter uses positive feature mappings to avoid explicitly constructing the $T \times T$ attention matrix, reducing the memory cost of conventional self-attention.

Cross-variable interactions are then inferred directly from the refined representations without explicit dependency graphs or message passing:

\begin{equation}
	C_{t,p,q}
	=
	\tanh
	\left(
	H_{t}^{\prime(p)\top}H_{t}^{\prime(q)}
	\right),
\end{equation}

yielding $C \in \mathbb{R}^{T \times P \times P}$. Thus, variable interactions are inferred dynamically from the current telemetry window rather than imposed by a predefined system topology.

\textbf{(5) Temporal Aggregation and Prediction.}

The refined representation $H'$ is aggregated across time using max pooling:

\begin{equation}
	Z^{(p)}
	=
	\max_t H_{t}^{\prime(p)}.
\end{equation}

This retains the strongest temporal evidence for each latent feature while avoiding recurrent state propagation. The resulting representation is passed to a lightweight prediction head:

\begin{equation}
	\hat{Y}
	=
	\psi_2(Z) + B,
\end{equation}

where $\psi_2(\cdot)$ denotes either a linear or shallow nonlinear head and $B$ is a learnable bias. The predicted representation is finally mapped back to the original measurement space using the inverse orthogonal transformation when applicable.
\subsection{Training Objective}

SFlexRCA jointly learns the expected system behavior and an explicit cross-variable interaction representation. The overall objective is

\begin{equation}
	\mathcal{L}
	=
	\mathcal{L}_{\mathrm{pred}}
	+
	\mathcal{L}_{\mathrm{reg}},
\end{equation}

where

\begin{equation}
	\mathcal{L}_{\mathrm{pred}}
	=
	\left|
	\hat{Y}_{t+1}-Y_{t+1}
	\right|_2^2
\end{equation}

measures the prediction error, and

\begin{equation}
	\mathcal{L}_{\mathrm{reg}}
	=
	\lambda_1\mathcal{L}_{\mathrm{sparse}}(C)
	+
	\lambda_2\mathcal{L}_{\mathrm{smooth}}(C)
	+
	\lambda_3\mathcal{L}_{\mathrm{KL}}(C)
\end{equation}

regularizes the learned interaction representation. 
The sparsity, temporal smoothness, and KL terms therefore directly constrain the cross-variable interactions inferred from the telemetry. This encourages the model to represent normal IIoT behavior using a compact and temporally consistent interaction structure while jointly optimizing predictive accuracy.

At inference time, SFlexRCA ranks monitored variables using their prediction residuals. The interaction representation is not treated as a separate fault score; instead, it is learned jointly with the predictor and structurally constrained during training. This follows the common principle in neural RCA methods~\cite{cLSTM,285-encoder-decoder} of learning a predictive model whose internal representations are constrained to capture meaningful system dependencies, while SFlexRCA differs in explicitly constructing these dependencies from shared latent representations rather than maintaining a separate predictor or parameterization for each variable.

\subsection{Complexity Analysis}
\begin{table}[t]
	\centering
	\caption{Complexity comparison of RCA approaches. $P$ is the number of monitored variables, $H$ the hidden dimension, $K$ the temporal order, and $L$ the number of hidden layers.}
	\label{tab:complexity}
	\small
	\begin{tabular}{p{1.7cm} p{2.0cm} p{3.0cm}}
		\toprule
		\textbf{Method} & \textbf{Complexity} & \textbf{Description} \\
		\midrule
		cMLP
		& $O(P^2KH)$
		& Separate MLP predictor for each target variable. \\
		\midrule
		cLSTM
		& $O(P^2KH+PH^2)$
		& Separate recurrent predictor for each target variable. \\
		\midrule
		GVAR
		& $O(P^2K)$
		& Explicitly models pairwise variable relationships across lags. \\
		\midrule
		AERCA
		& $O(3KHP^2)$
		& Three GVAR-based networks generate a full $P\times P$ matrix at each lag. \\
		\midrule
		CUTS+
		& $O(LPH^2+PH+P^2)$
		& Uses graph-based recurrent message passing with pairwise node interactions. \\
		\midrule
		SFlexRCA
		& $O(PH+H^2+HK)$
		& Shared lightweight model; avoids explicit pairwise coefficient matrices. \\
		\bottomrule
	\end{tabular}
\end{table}

The computational complexity of SFlexRCA can be summarized according to its five architectural stages. Steps (1) and (2) perform the orthogonal representation and shared variable encoding, with the latter mapping each variable into the shared hidden space, giving a dominant complexity of $O(PH)$. Steps (3) and (4) perform temporal context modeling and context refinement using shared latent representations, resulting in a complexity of $O(H^2)$ without introducing separate neural networks for individual variables. Step (5) performs temporal aggregation and prediction in the hidden space, contributing $O(HK)$. Consequently, the overall complexity is summarized as
\begin{equation}
	O(PH + H^2 + HK).
\end{equation}
This formulation highlights that SFlexRCA performs its main neural computation in a shared hidden space rather than replicating the computation across monitored variables. In contrast, cMLP and cLSTM introduce complexity that grows with the number of variables, GVAR and AERCA introduce multiplicative dependence on the temporal order through explicit coefficient modeling, and CUTS+ incurs additional complexity from graph-based message passing and its associated layers. SFlexRCA therefore shifts the dominant computation toward the compact hidden dimension $H$, while requiring only a shared linear dependence on the number of monitored variables. Table~\ref{tab:complexity} summarizes the complexity of SFlexRCA and representative causal RCA methods.

\section{Experiments}

We evaluate SFlexRCA on three IIoT datasets to assess its effectiveness for RCA under the computational constraints of edge deployment. We compare SFlexRCA with statistical and causal deep learning baselines and investigate: (1) RCA accuracy across multiple datasets, (2) training and inference efficiency, particularly on resource-constrained Raspberry Pi devices, (3) the contribution of architectural and loss components, and (4) sensitivity to the temporal window size.

\subsection{Datasets}

We evaluate SFlexRCA on three industrial control system datasets with different numbers of monitored variables and temporal characteristics, as summarized in Table~\ref{tab:dataset_comparison}. 
BATADAL~\cite{BATADAL} is a water distribution benchmark containing 43 monitored variables and captures temporal dynamics in a water distribution system.
SWaT~\cite{295-mathur2016-swat} is a water treatment benchmark with 51 monitored variables and longer temporal dependencies arising from the interactions among multiple physical processes.
WADI~\cite{wadi-34-of-301} is a water distribution benchmark containing 127 monitored variables and provides a higher-dimensional setting for evaluating the scalability of RCA methods.
Together, these datasets provide progressively different dimensional and temporal settings for evaluating root cause localization.

\begin{table}[t]
	\centering
	\caption{Comparison of the BATADAL, SWaT, and WADI datasets.}
	\label{tab:dataset_comparison}
	\scriptsize
	\begin{tabular}{p{2.2cm} p{1.5cm} p{1.5cm} p{1.5cm}}
		\toprule
		\textbf{Metric} & \textbf{BATADAL} & \textbf{SWaT} & \textbf{WADI} \\
		\midrule
		\textbf{System Type}
		& Simulated water distribution (C-Town)
		& Physical water treatment testbed
		& Physical water distribution testbed \\
		\midrule
		\textbf{Sampling Frequency}
		& 1 hour
		& 1 second
		& 1 second \\
		\midrule
		\textbf{Number of Channels}
		& 43 sensor/actuator features
		& 51 sensor/actuator features
		& 127 sensor/actuator features \\
		\midrule
		\textbf{Training Data (Normal)}
		& 8,760 samples (365 days)
		& 410,400--480,800 samples (7 days)
		& $\sim$1,200,000 samples (14 days) \\
		\midrule
		\textbf{Amount of Attacks}
		& 14 attacks total (7 train, 7 test)
		& 36--41 attack scenarios (4 days)
		& 15 attack scenarios (2 days) \\
		\bottomrule
	\end{tabular}
\end{table}
\subsection{Experimental Setup}

\paragraph{Baselines and Comparison.}

We compare SFlexRCA against statistical, non-causal, and causal deep learning RCA methods. Statistical baselines include RCD~\cite{286-Ikram2022-RCD}, which identifies influential variables using conditional independence testing, BARO~\cite{288-BARO}, which performs robust statistical root cause scoring, and TORAI~\cite{TORAI}, adapted to metric-only inputs for a consistent comparison. Non-causal baselines include TimeMixer++~\cite{timemixer++} and iTransformer~\cite{itransformer}, which model multivariate temporal dependencies without explicitly learning causal relationships. For causal deep learning, we consider per-variable architectures (cLSTM and cMLP~\cite{cLSTM}), lag-specific architectures (GVAR~\cite{GVAR} and AERCA~\cite{285-encoder-decoder}), and shared message-passing architectures (CUTS+~\cite{CUTS_PLUS}).
To assess the contribution of the proposed representation, we compare orthogonal projection with seven alternatives: no projection, learned projection, and static Legendre, Laguerre, Chebyshev, Hermite, and Fourier projections, following the experimental methodology of~\cite{280-FreqTimeLoss,olinear}.
All methods operate on the same metric-only telemetry inputs, and RCA scores are ranked using the same evaluation procedure. In particular, no additional anomaly thresholding or EVT-based post-processing is applied to SFlexRCA, ensuring a consistent comparison with the statistical and causal baselines.

\paragraph{Evaluation Metrics.}

We evaluate the methods from two complementary perspectives: RCA accuracy and computational efficiency.
For RCA accuracy, we report Recall at Top-$k$ (AC@k) for $k \in {1,5,10}$ and AUC@10, following established RCA evaluation protocols~\cite{286-Ikram2022-RCD,285-li2022-recall}. AC@$k$ measures whether the ground-truth root-cause variables are included among the top-$k$ variables ranked by each method, while AUC@10 evaluates the ranking quality across the top-$10$ predictions.
For training efficiency, we report the number of trainable parameters, peak memory consumption, and average training time per epoch. For inference efficiency, we report average inference latency per sample on resource-constrained edge devices. Specifically, experiments are conducted on a Raspberry Pi 5 equipped with an ARM Cortex-A76 processor and 16~GB RAM, and a Raspberry Pi 3 equipped with an ARM Cortex-A53 processor and 1~GB RAM. For edge energy evaluation, inference was performed on the Raspberry Pi 5 while monitoring PMIC power using the \texttt{vcgencmd pmic\_read\_adc} interface. Total energy was obtained by trapezoidal integration of the measured power trace over the inference workload, and energy per sample was calculated by dividing the measured energy by the total number of evaluated samples.

\paragraph{Implementation Details.}

All experiments are conducted using multiple random seeds on a Linux workstation equipped with an Intel(R) Core(TM) i9-10900K CPU @ 3.70GHz (20 cores), 32~GB RAM, and an NVIDIA GeForce RTX 3070 GPU with 8~GB memory, running Ubuntu 22.04.2 LTS. The implementation uses Python 3.10.12, PyTorch 2.7.1+cu126 with CUDA 12.6, and PyTorch Geometric 2.6.1.
Hyperparameters are tuned independently for each dataset, and the best configuration is selected. The latent dimension $D$ is set to 128 for SWaT and WADI and 256 for BATADAL. The number of training epochs is set to 200 for SWaT and BATADAL and 50 for WADI, with early stopping applied. The learning rate is set to $1\times10^{-4}$ across all experiments. For SFlexRCA, the orthogonal transformation is implemented using a dataset-specific linear projection.

\subsection{RQ1: Root Cause Analysis Accuracy}

\begin{table*}[t]
\caption{Root Cause Analysis Accuracy and Model Parameters Across BATADAL and SWAT and WADI (Mean $\pm$ SD).}
\label{tab:RQ1}
\centering
\scriptsize
\setlength{\tabcolsep}{2.0pt}
\renewcommand{\arraystretch}{0.92}
\setlength{\fboxsep}{1.0pt}
\begin{tabular}{l l *{4}{c} | l *{4}{c} | l *{4}{c}}
\toprule
 & \multicolumn{5}{c}{\textbf{BATADAL} \scriptsize{(K=16, P=43, training samples=8k)} } & \multicolumn{5}{c}{\textbf{SWAT} \scriptsize{(K=8, P=51, training samples=400k)}} & \multicolumn{5}{c}{\textbf{WADI} \scriptsize{(K=8, P=127, training samples=1.2M)}} \\
\cmidrule(lr){2-6}\cmidrule(lr){7-11}\cmidrule(lr){12-16}
\textbf{Method} & \textbf{Params} & \textbf{AC@1} & \textbf{AC@5} & \textbf{AC@10} & \textbf{auc@10} & \textbf{Params} & \textbf{AC@1} & \textbf{AC@5} & \textbf{AC@10} & \textbf{auc@10} & \textbf{Params} & \textbf{AC@1} & \textbf{AC@5} & \textbf{AC@10} & \textbf{auc@10} \\
\midrule
\multicolumn{12}{l}{\textbf{Statistical Methods}} \\
rcd & 0 & \cellcolor{yellow!0!red!35}0.000{\tiny$\pm$0.00} & \cellcolor{yellow!0!red!35}0.086{\tiny$\pm$0.00} & \cellcolor{yellow!0!red!35}0.070{\tiny$\pm$0.00} & \cellcolor{yellow!0!red!35}0.091{\tiny$\pm$0.00} & 0 & \cellcolor{yellow!0!red!35}0.000{\tiny$\pm$0.00} & \cellcolor{yellow!0!red!35}0.000{\tiny$\pm$0.00} & \cellcolor{yellow!35!red!35}0.300{\tiny$\pm$0.00} & \cellcolor{yellow!0!red!35}0.100{\tiny$\pm$0.00} & 0 & \cellcolor{yellow!0!red!35}0.000{\tiny$\pm$0.00} & \cellcolor{yellow!0!red!35}0.000{\tiny$\pm$0.00} & \cellcolor{yellow!0!red!35}0.000{\tiny$\pm$0.00} & \cellcolor{yellow!0!red!35}0.000{\tiny$\pm$0.00} \\
baro & 0 & \cellcolor{green!20!yellow!35}0.286{\tiny$\pm$0.00} & \cellcolor{green!60!yellow!35}0.162{\tiny$\pm$0.00} & \cellcolor{green!24!yellow!35}0.221{\tiny$\pm$0.00} & \cellcolor{green!45!yellow!35}0.200{\tiny$\pm$0.00} & 0 & \cellcolor{yellow!0!red!35}0.000{\tiny$\pm$0.00} & \cellcolor{green!9!yellow!35}0.200{\tiny$\pm$0.00} & \cellcolor{yellow!0!red!35}0.250{\tiny$\pm$0.00} & \cellcolor{yellow!64!red!35}0.180{\tiny$\pm$0.00} & 0 & \cellcolor{green!59!yellow!35}0.267{\tiny$\pm$0.00} & \cellcolor{green!84!yellow!35}0.267{\tiny$\pm$0.00} & \cellcolor{green!88!yellow!35}0.356{\tiny$\pm$0.00} & \cellcolor{green!82!yellow!35}0.283{\tiny$\pm$0.00} \\
torai & 0 & \cellcolor{green!20!yellow!35}0.286{\tiny$\pm$0.00} & \cellcolor{green!93!yellow!35}0.178{\tiny$\pm$0.03} & \cellcolor{green!100!yellow!35}0.312{\tiny$\pm$0.04} & \cellcolor{green!100!yellow!35}0.241{\tiny$\pm$0.03} & 0 & \cellcolor{yellow!77!red!35}0.117{\tiny$\pm$0.03} & \cellcolor{green!9!yellow!35}0.200{\tiny$\pm$0.00} & \cellcolor{yellow!70!red!35}0.350{\tiny$\pm$0.00} & \cellcolor{green!2!yellow!35}0.227{\tiny$\pm$0.01} & 0 & \cellcolor{green!59!yellow!35}0.267{\tiny$\pm$0.00} & \cellcolor{green!84!yellow!35}0.267{\tiny$\pm$0.00} & \cellcolor{green!47!yellow!35}0.278{\tiny$\pm$0.00} & \cellcolor{green!73!yellow!35}0.269{\tiny$\pm$0.00} \\
\midrule
\multicolumn{12}{l}{\textbf{Non-Causal Methods}} \\
TimeMixer++ & 576.6K & \cellcolor{yellow!39!red!35}0.095{\tiny$\pm$0.08} & \cellcolor{yellow!73!red!35}0.121{\tiny$\pm$0.06} & \cellcolor{green!2!yellow!35}0.194{\tiny$\pm$0.06} & \cellcolor{yellow!45!red!35}0.125{\tiny$\pm$0.07} & 160.5K & \cellcolor{green!11!yellow!35}0.167{\tiny$\pm$0.03} & \cellcolor{green!45!yellow!35}0.267{\tiny$\pm$0.03} & \cellcolor{yellow!82!red!35}0.367{\tiny$\pm$0.04} & \cellcolor{green!39!yellow!35}0.272{\tiny$\pm$0.04} & 199.9K & \cellcolor{green!59!yellow!35}0.267{\tiny$\pm$0.00} & \cellcolor{green!110!yellow!35}0.304{\tiny$\pm$0.06} & \cellcolor{green!72!yellow!35}0.326{\tiny$\pm$0.04} & \cellcolor{green!95!yellow!35}0.302{\tiny$\pm$0.05} \\
iTransformer & 800.0K & \cellcolor{yellow!19!red!35}0.048{\tiny$\pm$0.08} & \cellcolor{yellow!-26!red!35}0.073{\tiny$\pm$0.04} & \cellcolor{yellow!90!red!35}0.180{\tiny$\pm$0.03} & \cellcolor{yellow!6!red!35}0.096{\tiny$\pm$0.04} & 201.4K & \cellcolor{green!11!yellow!35}0.167{\tiny$\pm$0.03} & \cellcolor{green!36!yellow!35}0.250{\tiny$\pm$0.09} & \cellcolor{green!11!yellow!35}0.408{\tiny$\pm$0.10} & \cellcolor{green!46!yellow!35}0.281{\tiny$\pm$0.07} & 201.4K & \cellcolor{green!126!yellow!35}0.378{\tiny$\pm$0.04} & \cellcolor{green!138!yellow!35}0.344{\tiny$\pm$0.04} & \cellcolor{green!96!yellow!35}0.370{\tiny$\pm$0.01} & \cellcolor{green!128!yellow!35}0.354{\tiny$\pm$0.03} \\
\midrule
\multicolumn{12}{l}{\textbf{Causal Methods}} \\
cMLP & 7.1M & \cellcolor{green!20!yellow!35}0.286{\tiny$\pm$0.00} & \cellcolor{yellow!60!red!35}0.114{\tiny$\pm$0.00} & \cellcolor{yellow!59!red!35}0.142{\tiny$\pm$0.01} & \cellcolor{yellow!70!red!35}0.144{\tiny$\pm$0.01} & 2.3M & \cellcolor{yellow!66!red!35}0.100{\tiny$\pm$0.00} & \cellcolor{green!63!yellow!35}0.300{\tiny$\pm$0.05} & \cellcolor{green!100!yellow!35}0.533{\tiny$\pm$0.03} & \cellcolor{green!85!yellow!35}0.328{\tiny$\pm$0.03} & 14.5M & \cellcolor{green!59!yellow!35}0.267{\tiny$\pm$0.00} & \cellcolor{green!84!yellow!35}0.267{\tiny$\pm$0.00} & \cellcolor{green!52!yellow!35}0.289{\tiny$\pm$0.04} & \cellcolor{green!77!yellow!35}0.276{\tiny$\pm$0.02} \\
cLSTM & 13.3M & \cellcolor{green!60!yellow!35}0.381{\tiny$\pm$0.08} & \cellcolor{green!99!yellow!35}0.181{\tiny$\pm$0.04} & \cellcolor{yellow!99!red!35}0.190{\tiny$\pm$0.05} & \cellcolor{green!67!yellow!35}0.216{\tiny$\pm$0.05} & 4.7M & \cellcolor{green!100!yellow!35}0.300{\tiny$\pm$0.00} & \cellcolor{green!81!yellow!35}0.333{\tiny$\pm$0.03} & \cellcolor{yellow!94!red!35}0.383{\tiny$\pm$0.06} & \cellcolor{green!91!yellow!35}0.337{\tiny$\pm$0.03} & 16.7M & \cellcolor{green!86!yellow!35}0.311{\tiny$\pm$0.04} & \cellcolor{green!87!yellow!35}0.271{\tiny$\pm$0.01} & \cellcolor{green!60!yellow!35}0.304{\tiny$\pm$0.03} & \cellcolor{green!79!yellow!35}0.279{\tiny$\pm$0.00} \\
GVAR & 14.2M & \cellcolor{yellow!60!red!35}0.143{\tiny$\pm$0.14} & \cellcolor{green!100!yellow!35}0.181{\tiny$\pm$0.04} & \cellcolor{green!57!yellow!35}0.261{\tiny$\pm$0.03} & \cellcolor{green!41!yellow!35}0.197{\tiny$\pm$0.05} & 3.2M & \cellcolor{green!88!yellow!35}0.283{\tiny$\pm$0.03} & \cellcolor{green!90!yellow!35}0.350{\tiny$\pm$0.00} & \cellcolor{yellow!88!red!35}0.375{\tiny$\pm$0.03} & \cellcolor{green!94!yellow!35}0.340{\tiny$\pm$0.01} & 14.8M & \cellcolor{yellow!39!red!35}0.067{\tiny$\pm$0.00} & \cellcolor{yellow!84!red!35}0.122{\tiny$\pm$0.10} & \cellcolor{yellow!72!red!35}0.137{\tiny$\pm$0.12} & \cellcolor{yellow!68!red!35}0.107{\tiny$\pm$0.08} \\
AERCA & 42.6M & \cellcolor{green!20!yellow!35}0.286{\tiny$\pm$0.00} & \cellcolor{green!19!yellow!35}0.143{\tiny$\pm$0.03} & \cellcolor{yellow!54!red!35}0.137{\tiny$\pm$0.04} & \cellcolor{yellow!73!red!35}0.146{\tiny$\pm$0.02} & 9.6M & \cellcolor{yellow!100!red!35}0.150{\tiny$\pm$0.00} & \cellcolor{green!72!yellow!35}0.317{\tiny$\pm$0.03} & \cellcolor{green!41!yellow!35}0.450{\tiny$\pm$0.05} & \cellcolor{green!93!yellow!35}0.338{\tiny$\pm$0.03} & 44.4M & \cellcolor{yellow!39!red!35}0.067{\tiny$\pm$0.00} & \cellcolor{yellow!41!red!35}0.060{\tiny$\pm$0.05} & \cellcolor{yellow!31!red!35}0.059{\tiny$\pm$0.04} & \cellcolor{yellow!39!red!35}0.061{\tiny$\pm$0.04} \\
CUTS+ & 265.8K & \cellcolor{yellow!60!red!35}0.143{\tiny$\pm$0.14} & \cellcolor{green!39!yellow!35}0.152{\tiny$\pm$0.07} & \cellcolor{green!58!yellow!35}0.262{\tiny$\pm$0.06} & \cellcolor{green!28!yellow!35}0.187{\tiny$\pm$0.05} & 91.6K & \cellcolor{green!44!yellow!35}0.217{\tiny$\pm$0.03} & \cellcolor{green!63!yellow!35}0.300{\tiny$\pm$0.05} & \cellcolor{yellow!64!red!35}0.342{\tiny$\pm$0.12} & \cellcolor{green!51!yellow!35}0.287{\tiny$\pm$0.06} & 163.6K & \cellcolor{yellow!79!red!35}0.133{\tiny$\pm$0.00} & \cellcolor{green!69!yellow!35}0.244{\tiny$\pm$0.02} & \cellcolor{green!47!yellow!35}0.278{\tiny$\pm$0.04} & \cellcolor{green!52!yellow!35}0.237{\tiny$\pm$0.01} \\
\midrule
\multicolumn{12}{l}{\textbf{Proposed Method}} \\
SFlexRCA & 279.4K & \cellcolor{green!100!yellow!35}0.476{\tiny$\pm$0.08} & \cellcolor{green!59!yellow!35}0.162{\tiny$\pm$0.02} & \cellcolor{green!18!yellow!35}0.213{\tiny$\pm$0.01} & \cellcolor{green!91!yellow!35}0.234{\tiny$\pm$0.01} & 108.1K & \cellcolor{green!55!yellow!35}0.233{\tiny$\pm$0.03} & \cellcolor{green!100!yellow!35}0.367{\tiny$\pm$0.08} & \cellcolor{green!17!yellow!35}0.417{\tiny$\pm$0.06} & \cellcolor{green!100!yellow!35}0.347{\tiny$\pm$0.04} & 35.3K & \cellcolor{green!100!yellow!35}0.333{\tiny$\pm$0.00} & \cellcolor{green!100!yellow!35}0.289{\tiny$\pm$0.02} & \cellcolor{green!100!yellow!35}0.378{\tiny$\pm$0.02} & \cellcolor{green!100!yellow!35}0.310{\tiny$\pm$0.01} \\
\bottomrule
\end{tabular}
\end{table*}

\textbf{RCA Accuracy Across Statistical and Learned Methods.}

Table~\ref{tab:RQ1} compares SFlexRCA with statistical, non-causal forecasting, and causal RCA methods across BATADAL, SWAT, and WADI. Statistical methods require no trainable parameters but provide inconsistent localization accuracy across datasets, reflecting their limited ability to capture complex temporal and cross-variable dependencies.

\textbf{Effect of Training Data and Causal Modeling}
Non-causal forecasting models, including TimeMixer++ and iTransformer, perform substantially better on SWAT and WADI than on BATADAL. This difference is particularly pronounced for BATADAL, where only approximately 8K normal training samples are available: iTransformer achieves an AC@1 of only 0.048, while TimeMixer++ achieves 0.095. With substantially larger training sets of approximately 400K and 1.2M samples on SWAT and WADI, respectively, both models become considerably more competitive, with iTransformer reaching AC@1 values of 0.167 and 0.378. These results suggest that forecasting-based approaches benefit substantially from abundant normal-operation data, but their localization performance can deteriorate under data-limited conditions. For IIoT environments where collecting large volumes of representative normal-operation data may be impractical, this dependence can limit their applicability.
In contrast, causal methods explicitly model temporal and cross-variable dependencies associated with system behavior, providing a mechanism more directly aligned with root-cause localization than forecasting alone. This provides an advantage when training data is limited. For example, cLSTM achieves an AC@1 of 0.381 on BATADAL, substantially exceeding the 0.048 and 0.095 obtained by iTransformer and TimeMixer++, respectively, despite all methods being trained using the same limited dataset. However, this dependency modeling generally comes at a substantial architectural cost, with parameter requirements increasing considerably as the number of monitored variables and temporal dependencies grow. On WADI, for example, cMLP and cLSTM require 14.5M and 16.7M parameters, respectively, while GVAR and AERCA require 14.8M and 44.4M parameters.

\textbf{RCA Accuracy with Compact Architecture}
SFlexRCA achieves competitive or superior RCA accuracy while maintaining a substantially smaller architecture than most causal deep learning baselines. On BATADAL, SFlexRCA achieves the highest AC@1 of 0.476 with only 279.4K parameters, compared with 13.3M for cLSTM and 7.1M for cMLP. The advantage is also seen on WADI, where SFlexRCA achieves AC@1, AC@5, and AC@10 of 0.333, 0.289, and 0.378, respectively, while requiring only 35.3K parameters. Notably, this lower parameter count on WADI does not result from having fewer monitored variables: WADI contains substantially more variables than both BATADAL and SWAT. Rather, as discussed in RQ3, the WADI configuration does not require the linear-attention component used in the other configurations, further reducing the model size. Thus, the increase in the number of monitored variables does not necessarily translate into increased model complexity in SFlexRCA. This scalability advantage stems from SFlexRCA's shared variable encoding, which avoids the per-variable neural architectures used by cMLP and cLSTM and therefore prevents the substantial parameter growth observed as the number of monitored variables increases. Similarly, GVAR and AERCA incur substantially larger models due to their dependency- and lag-specific architectures.
On SWAT, SFlexRCA remains competitive, achieving an AC@5 of 0.367, higher than the strongest baseline value of 0.350, while using only 108.1K parameters. Compared with the recent CUTS+ framework, SFlexRCA further achieves higher AC@1, AC@5, and AC@10 on SWAT, with values of 0.233, 0.367, and 0.417 compared with 0.217, 0.300, and 0.342, respectively. Although CUTS+ uses fewer parameters than SFlexRCA on BATADAL and SWAT, its message-passing architecture introduces additional computational overhead that is not reflected by parameter count alone, as examined in RQ2.

\textbf{Summary.}
Overall, these results highlight the central trade-off addressed by SFlexRCA: non-causal forecasting methods can become competitive when large volumes of normal-operation data are available, while causal methods can provide stronger dependency-aware localization but typically require substantially larger architectures. SFlexRCA combines dependency-aware modeling with a compact shared representation, achieving competitive or superior RCA accuracy with substantially fewer parameters, without requiring the per-variable, lag-specific, or explicit message-passing mechanisms of competing approaches, and while remaining effective in data-limited settings. The resulting training and inference efficiency are examined in RQ2.

\subsection{RQ-2: Efficiency Analysis}

\begin{figure}[t]
    \centering
    
    \begin{subfigure}[b]{0.235\textwidth}
        \centering
        \includegraphics[width=\linewidth]{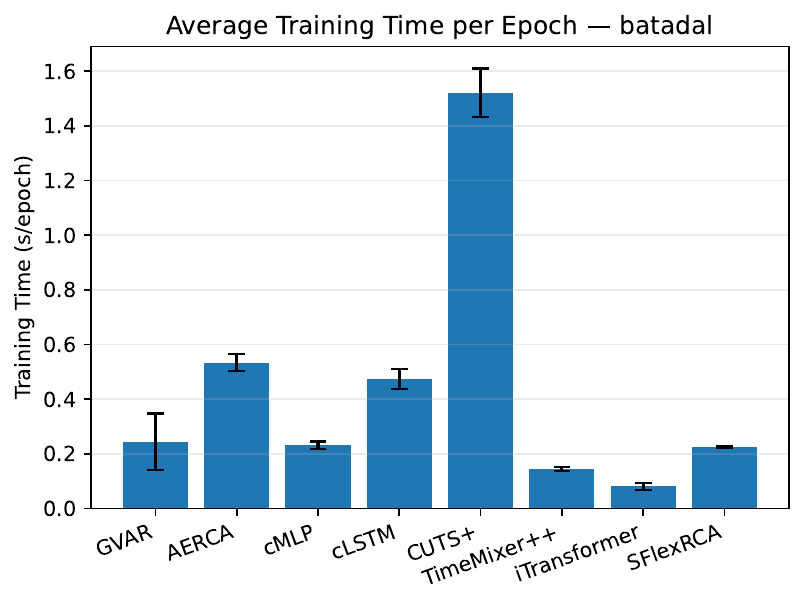}
        \caption{Avg training time per epoch (BATADAL).}
        \label{fig:rq2_batadal_training_time}
    \end{subfigure}
    \hfill
    \begin{subfigure}[b]{0.235\textwidth}
        \centering
        \includegraphics[width=\linewidth]{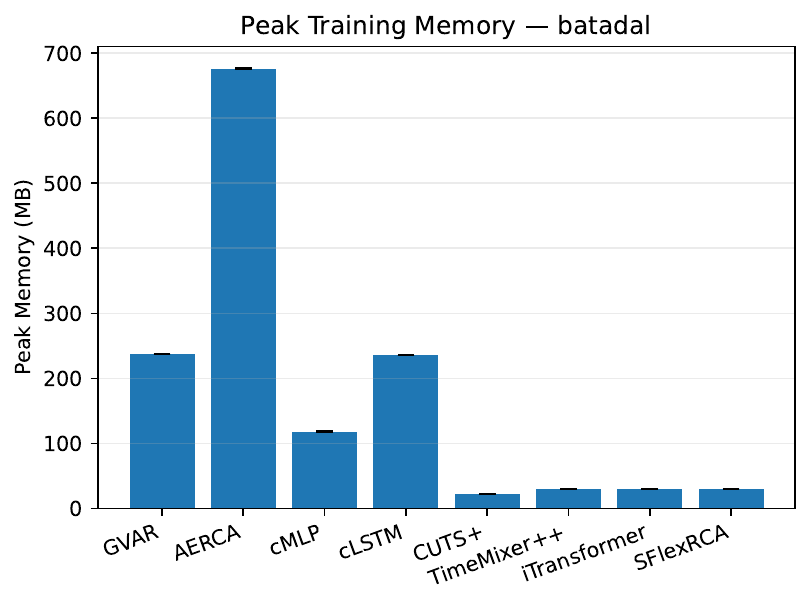}
        \caption{Peak training memory (BATADAL).}
        \label{fig:rq2_batadal_memory}
    \end{subfigure}

    \vspace{0.5em}

    \begin{subfigure}[b]{0.235\textwidth}
        \centering
        \includegraphics[width=\linewidth]{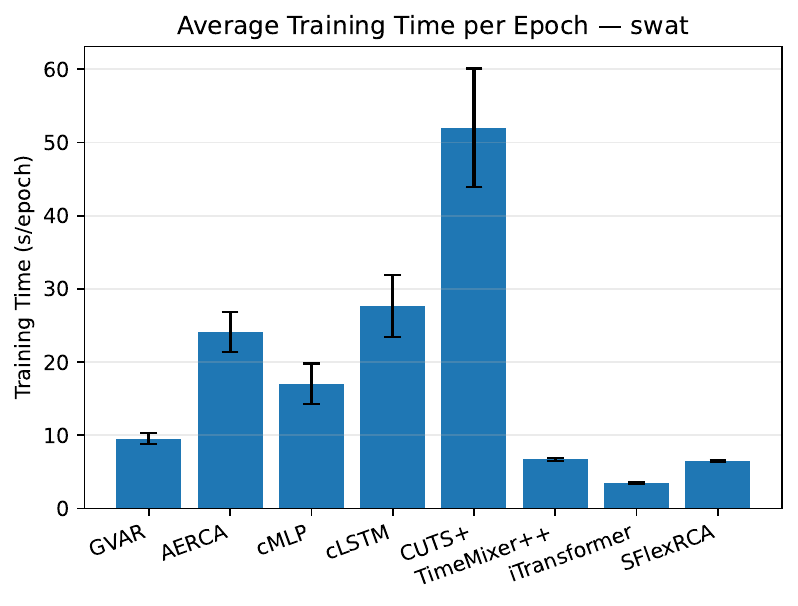}
        \caption{Avg training time per epoch (SWaT).}
        \label{fig:rq2_swat_training_time}
    \end{subfigure}
    \hfill
    \begin{subfigure}[b]{0.235\textwidth}
        \centering
        \includegraphics[width=\linewidth]{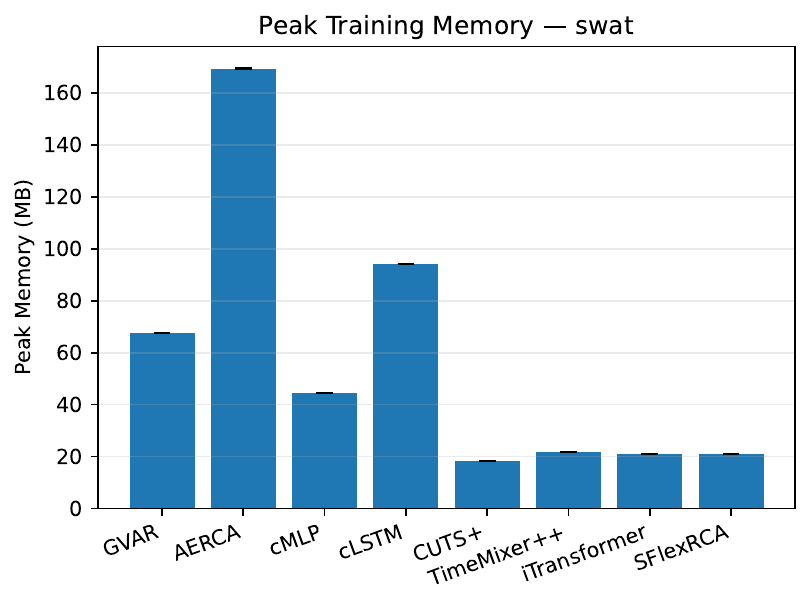}
        \caption{Peak training memory (SWaT).}
        \label{fig:rq2_swat_memory}
    \end{subfigure}

    \vspace{0.5em}

    \begin{subfigure}[b]{0.235\textwidth}
        \centering
        \includegraphics[width=\linewidth]{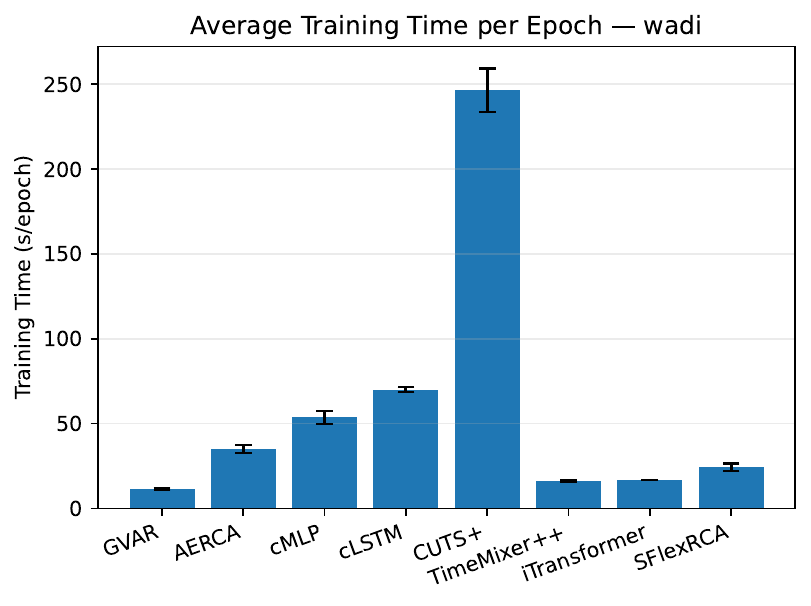}
        \caption{Avg training time per epoch (WADI).}
        \label{fig:rq2_wadi_training_time}
    \end{subfigure}
    \hfill
    \begin{subfigure}[b]{0.235\textwidth}
        \centering
        \includegraphics[width=\linewidth]{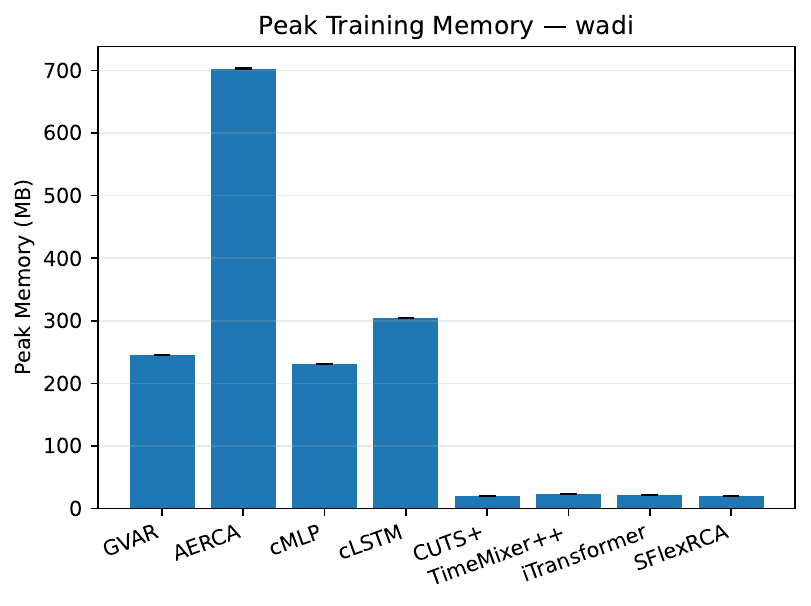}
        \caption{Peak training memory (WADI).}
        \label{fig:rq2_wadi_memory}
    \end{subfigure}

    \caption{Training efficiency metrics (average training time per epoch and peak training memory) across BATADAL, SWaT, and WADI datasets.}
    \label{fig:rq2_training_efficiency}
\end{figure}
\begin{figure*}[t]
    \centering
    \begin{subfigure}[b]{0.235\textwidth}
        \centering
        \includegraphics[width=\linewidth]{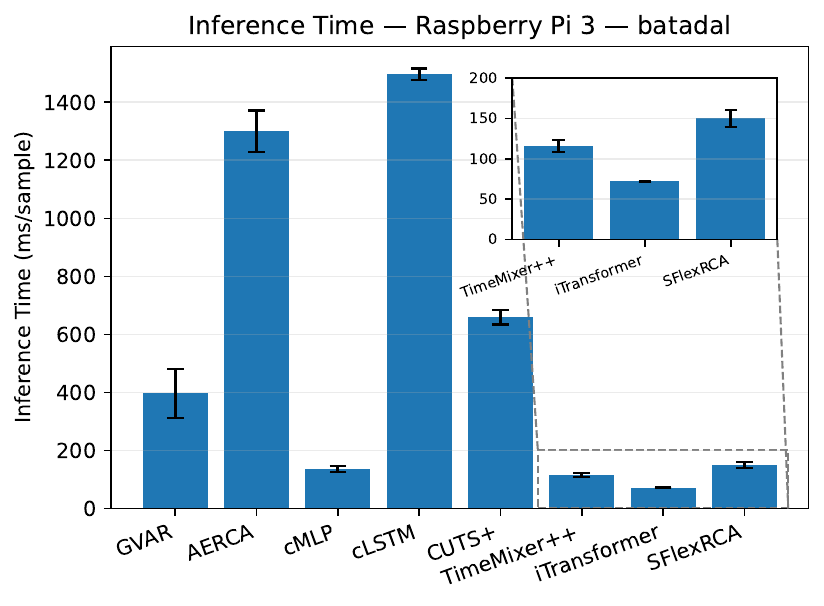}
        \caption{Inference time: Pi 3 (BATADAL).}
        \label{fig:rq2_batadal_pi3_time}
    \end{subfigure}
    \hfill
    \begin{subfigure}[b]{0.235\textwidth}
        \centering
        \includegraphics[width=\linewidth]{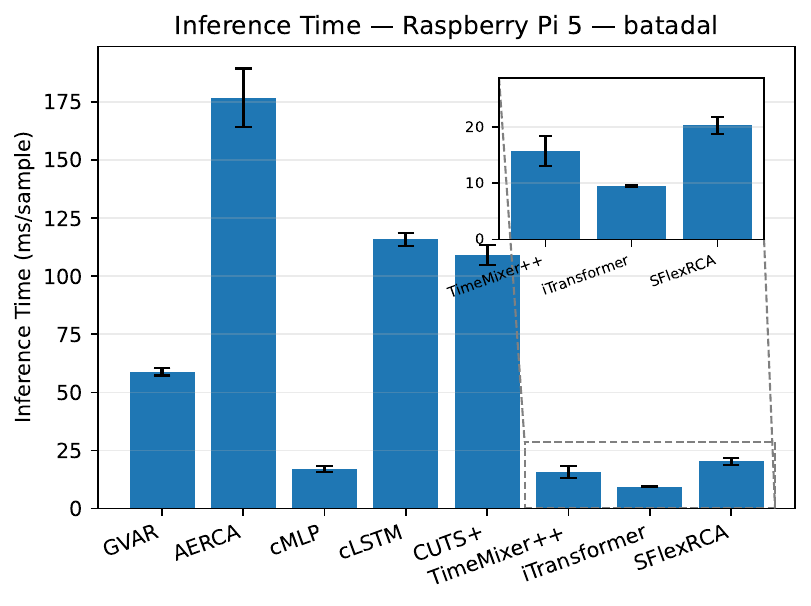}
        \caption{Inference time: Pi 5 (BATADAL).}
        \label{fig:rq2_batadal_pi5_time}
    \end{subfigure}
    \hfill
    \begin{subfigure}[b]{0.235\textwidth}
        \centering
        \includegraphics[width=\linewidth]{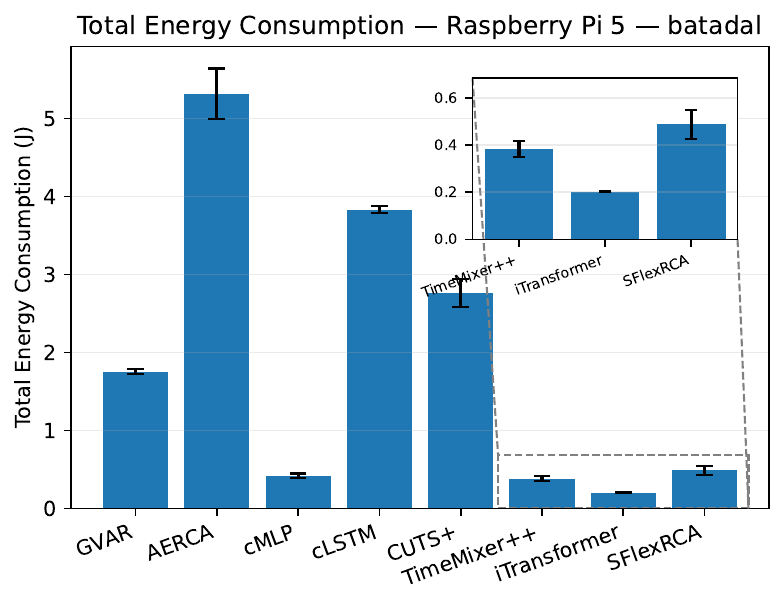}
        \caption{Total energy: Pi 5 (BATADAL).}
        \label{fig:rq2_batadal_pi5_total_energy}
    \end{subfigure}
    \hfill
    \begin{subfigure}[b]{0.235\textwidth}
        \centering
        \includegraphics[width=\linewidth]{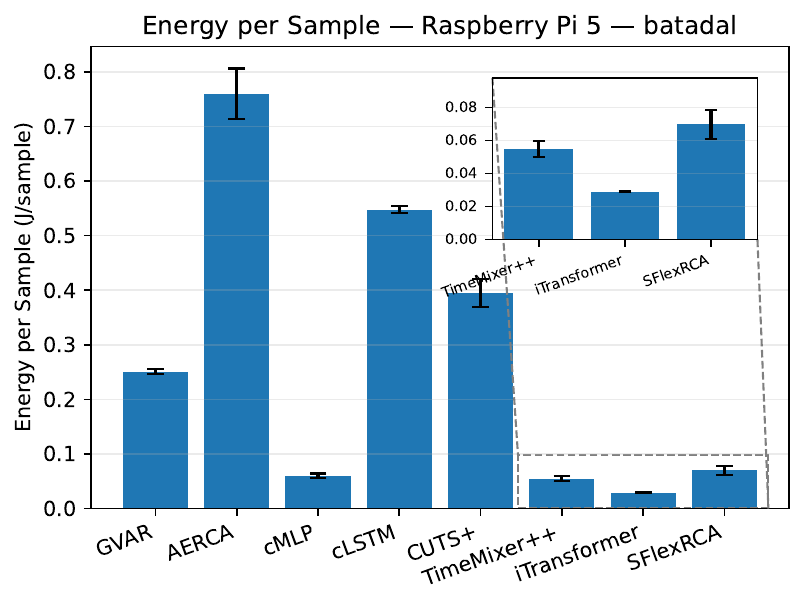}
        \caption{Energy / sample: Pi 5 (BATADAL).}
        \label{fig:rq2_batadal_pi5_energy_per_sample}
    \end{subfigure}

    \vspace{0.5em}

    \begin{subfigure}[b]{0.235\textwidth}
        \centering
        \includegraphics[width=\linewidth]{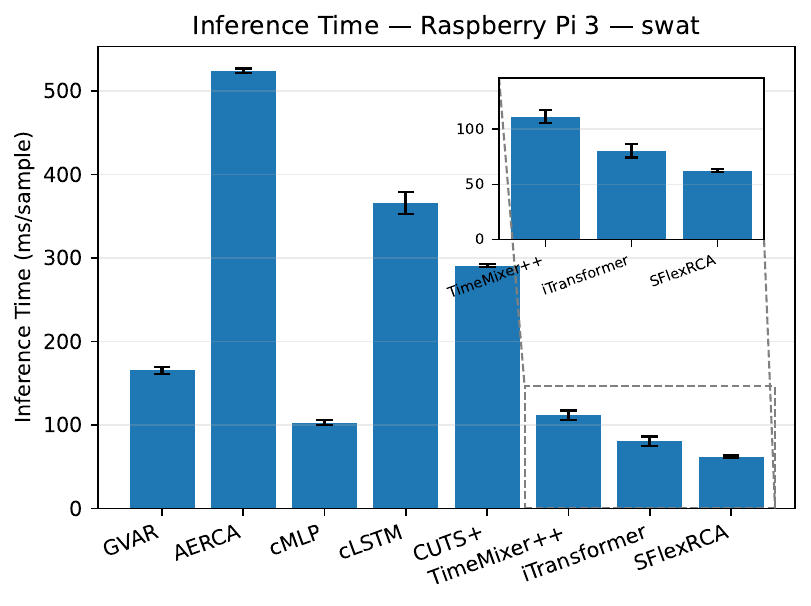}
        \caption{Inference time: Pi 3 (SWaT).}
        \label{fig:rq2_swat_pi3_time}
    \end{subfigure}
    \hfill
    \begin{subfigure}[b]{0.235\textwidth}
        \centering
        \includegraphics[width=\linewidth]{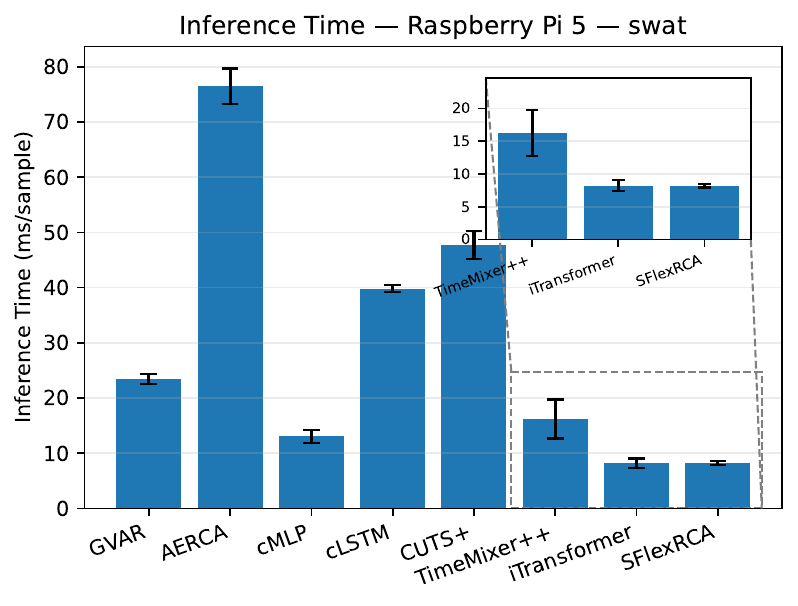}
        \caption{Inference time: Pi 5 (SWaT).}
        \label{fig:rq2_swat_pi5_time}
    \end{subfigure}
    \hfill
    \begin{subfigure}[b]{0.235\textwidth}
        \centering
        \includegraphics[width=\linewidth]{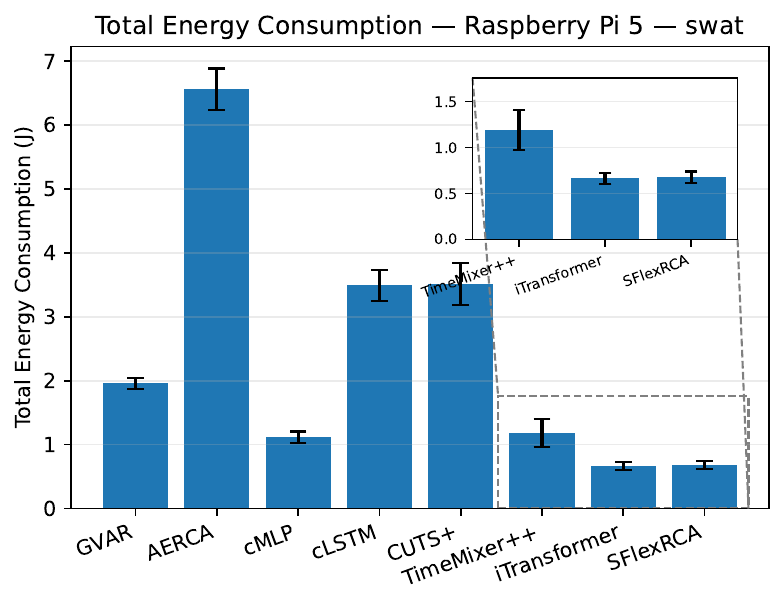}
        \caption{Total energy: Pi 5 (SWaT).}
        \label{fig:rq2_swat_pi5_total_energy}
    \end{subfigure}
    \hfill
    \begin{subfigure}[b]{0.235\textwidth}
        \centering
        \includegraphics[width=\linewidth]{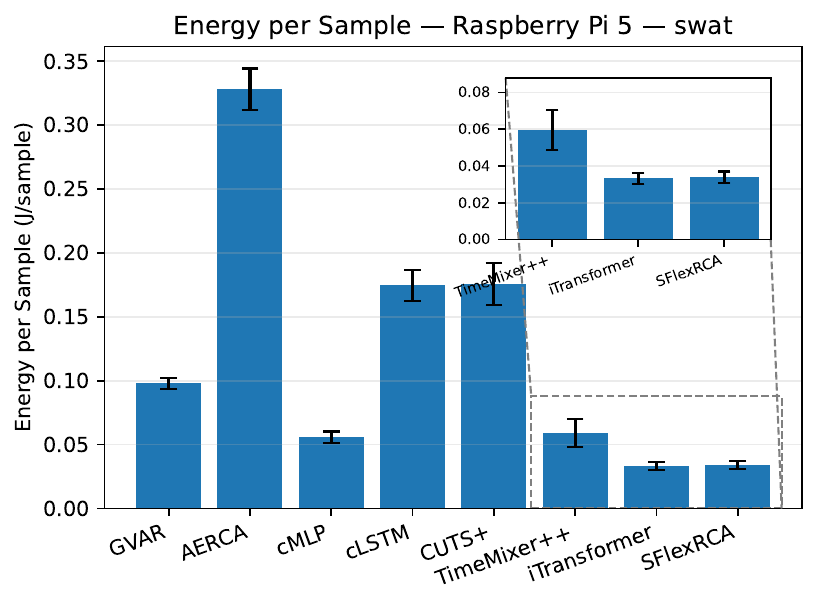}
        \caption{Energy / sample: Pi 5 (SWaT).}
        \label{fig:rq2_swat_pi5_energy_per_sample}
    \end{subfigure}

    \vspace{0.5em}

    \begin{subfigure}[b]{0.235\textwidth}
        \centering
        \includegraphics[width=\linewidth]{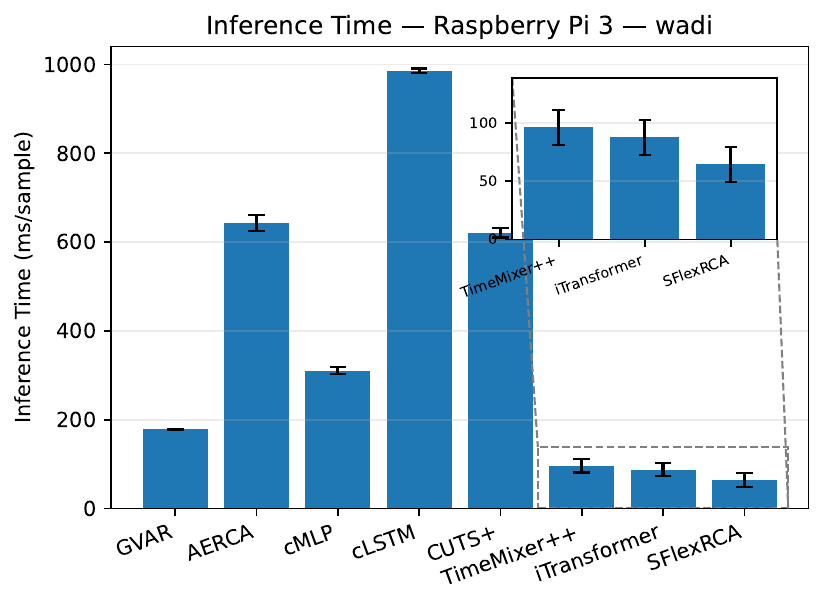}
        \caption{Inference time: Pi 3 (WADI).}
        \label{fig:rq2_wadi_pi3_time}
    \end{subfigure}
    \hfill
    \begin{subfigure}[b]{0.235\textwidth}
        \centering
        \includegraphics[width=\linewidth]{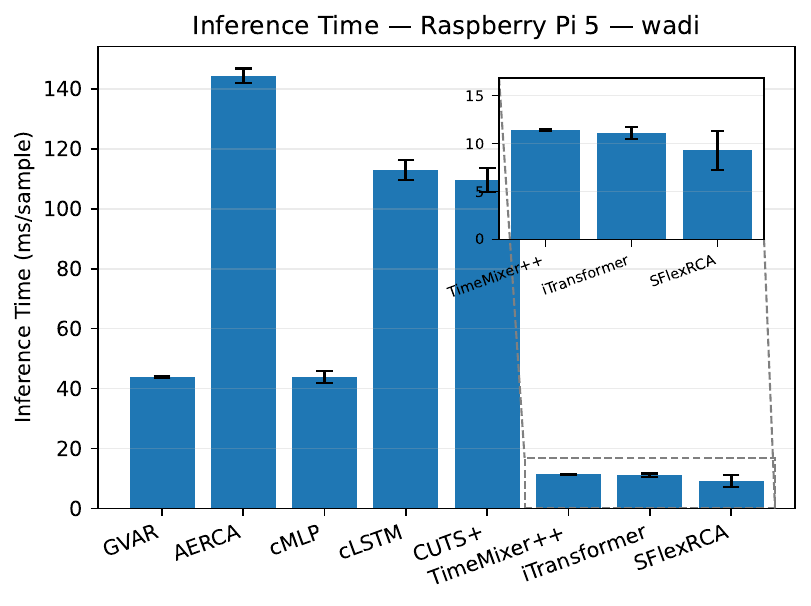}
        \caption{Inference time: Pi 5 (WADI).}
        \label{fig:rq2_wadi_pi5_time}
    \end{subfigure}
    \hfill
    \begin{subfigure}[b]{0.235\textwidth}
        \centering
        \includegraphics[width=\linewidth]{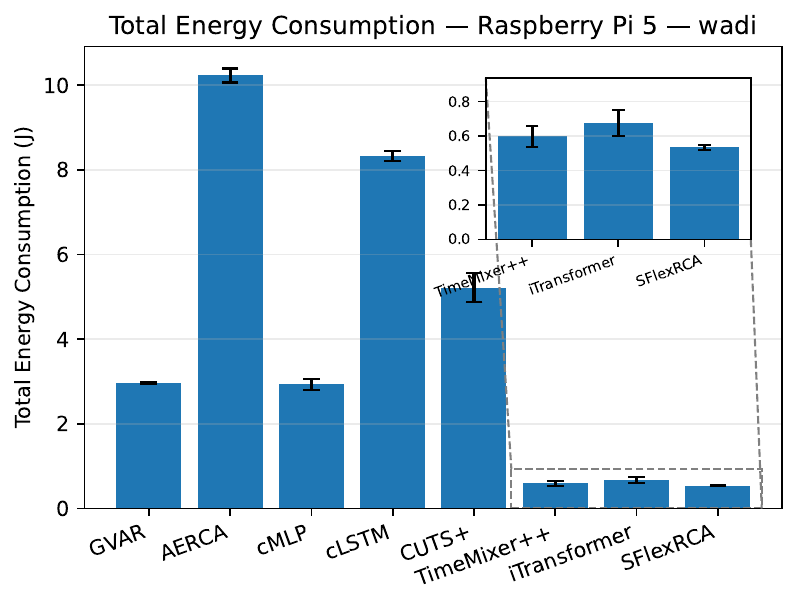}
        \caption{Total energy: Pi 5 (WADI).}
        \label{fig:rq2_wadi_pi5_total_energy}
    \end{subfigure}
    \hfill
    \begin{subfigure}[b]{0.235\textwidth}
        \centering
        \includegraphics[width=\linewidth]{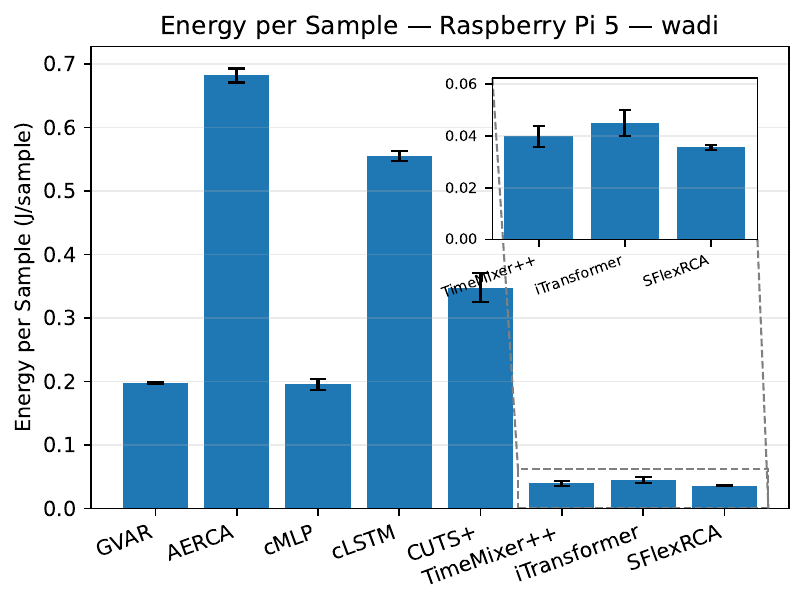}
        \caption{Energy / sample: Pi 5 (WADI).}
        \label{fig:rq2_wadi_pi5_energy_per_sample}
    \end{subfigure}

    \caption{Edge deployment efficiency metrics (inference latency on Pi 3/5 and energy consumption on Pi 5) across BATADAL, SWaT, and WADI datasets.}
    \label{fig:rq2_inference_efficiency}
\end{figure*}

In this section, we evaluate the training and inference efficiency of SFlexRCA compared with causal deep learning baselines. These measurements characterize the training and deployment costs of the evaluated RCA methods under realistic edge-device constraints.

\textbf{Training Efficiency}

SFlexRCA exhibits low training overhead across all three datasets in terms of both training time and peak memory consumption. On BATADAL, SWAT, and WADI, SFlexRCA requires 0.224, 6.51, and 24.24~s/epoch, respectively, while peak memory remains below 31~MB. Compared with the causal RCA methods, SFlexRCA generally provides lower training overhead. This is particularly evident on SWAT, where SFlexRCA requires 6.51~s/epoch, compared with 9.56~s for GVAR, 24.15~s for AERCA, 17.05~s for cMLP, 27.70~s for cLSTM, and 51.99~s for CUTS+. Its peak memory consumption is only 20.86~MB, substantially lower than GVAR (67.75~MB), AERCA (169.46~MB), cMLP (44.41~MB), and cLSTM (94.25~MB). Although CUTS+ uses only 18.15~MB, its message-passing architecture incurs considerably higher training time.
The non-causal models, TimeMixer++ and iTransformer, achieve the lowest training overhead overall. On SWAT, however, SFlexRCA remains highly competitive with these models, requiring 6.51~s/epoch compared with 6.57~s for TimeMixer++, while using slightly less memory (20.86~MB versus 21.71~MB). iTransformer achieves lower training time at 3.42~s/epoch, with a comparable memory footprint of 20.88~MB. Nevertheless, the forecasting-based formulation of these non-causal models typically benefits from sufficient normal-operation data, which may be limited in IIoT deployments. This limitation is particularly relevant to BATADAL, where observations are sampled at hourly intervals. As shown in RQ1, SFlexRCA can provide more effective RCA under such data-limited settings while retaining a compact training footprint. Overall, SFlexRCA provides a favorable balance between training efficiency and RCA effectiveness, combining the low resource requirements of non-causal models with substantially lower training overhead than most causal RCA architectures.

\textbf{Edge Inference Efficiency}
Inference experiments further demonstrate the suitability of SFlexRCA for resource-constrained IIoT edge devices. As expected, the Raspberry Pi~3 consistently incurs higher latency than the Raspberry Pi~5. On the Pi~3, SFlexRCA requires 149.8, 62.3, and 64.3~ms/sample on BATADAL, SWAT, and WADI, respectively. It achieves the lowest latency on SWAT and WADI, with the advantage becoming particularly pronounced on WADI, where it is substantially faster than GVAR (178.4~ms), cMLP (310.2~ms), CUTS+ (620.7~ms), AERCA (642.2~ms), and cLSTM (986.0~ms). On the Pi~5, SFlexRCA requires only 20.3, 8.2, and 9.3~ms/sample on BATADAL, SWAT, and WADI, respectively. In particular, SFlexRCA achieves the lowest latency among the causal RCA methods on all three datasets and is the overall fastest method on WADI.
The non-causal transformer models provide a useful comparison. On the Raspberry Pi~3, SFlexRCA achieves lower inference latency than both transformer models on SWAT and WADI, requiring 62.3 and 64.3~ms/sample compared with 80.3 and 87.3~ms/sample for iTransformer, respectively. On BATADAL, iTransformer is faster, requiring 71.8~ms/sample compared with 149.8~ms/sample for SFlexRCA. Similar behavior is observed on the Pi~5: iTransformer achieves the lowest latency on BATADAL (9.5~ms/sample) and is marginally faster than SFlexRCA on SWAT, whereas SFlexRCA becomes the fastest method on WADI, requiring only 9.3~ms/sample compared with 11.1~ms for iTransformer and 11.4~ms for TimeMixer++. These results indicate that SFlexRCA remains highly competitive with lightweight transformer architectures while providing more accurate RCA for systems with limited normal-operation data, as shown in RQ1.

\textbf{Edge Energy Consumption}
Energy consumption is evaluated on the Raspberry Pi~5, where reliable power measurements are available. SFlexRCA consumes 0.488, 0.678, and 0.534~J per evaluation on BATADAL, SWAT, and WADI, respectively, corresponding to 0.070, 0.034, and 0.036~J/sample. SFlexRCA achieves the lowest energy consumption on WADI (0.036~J/sample), while on SWAT it is essentially tied with iTransformer (0.034 versus 0.033~J/sample). On BATADAL, the non-causal transformer models achieve lower energy consumption, with iTransformer requiring only 0.029~J/sample and TimeMixer++ 0.055~J/sample, compared with 0.070~J/sample for SFlexRCA. This advantage is partly attributable to BATADAL having the smallest number of monitored sensors ($P=43$), which reduces the computational cost of the transformer-based models. However, as shown in RQ1, these lower energy costs do not translate into better RCA performance. Thus, SFlexRCA provides a favorable accuracy--efficiency trade-off across both causal and non-causal methods, placing it on the Pareto frontier rather than optimizing efficiency at the expense of RCA effectiveness. This trade-off is particularly relevant for continuous IIoT deployment, where both diagnostic accuracy and energy consumption accumulate over repeated inference.

\textbf{Summary.}
Overall, the results show that SFlexRCA provides a favorable accuracy--efficiency trade-off for edge-based IIoT RCA. Rather than optimizing only model size or individual inference latency, SFlexRCA maintains low training and inference overhead while preserving effective RCA across systems with substantially different numbers of monitored variables (BATADAL: 43, SWaT: 51, WADI: 127) and temporal characteristics (Table~\ref{tab:dataset_comparison}). Although non-causal transformer models can achieve lower computational and energy costs on smaller systems such as BATADAL, their lower RCA accuracy limits this efficiency advantage in practice. In contrast, SFlexRCA remains competitive with these efficient models while substantially reducing the overhead of most causal architectures. This balance places SFlexRCA on the accuracy--efficiency Pareto frontier across the evaluated methods and supports its use for continuous, on-device RCA in resource-constrained IIoT environments.

\subsection{RQ-3: Ablation Study}

\begin{table*}[t]
\caption{Case-Based Ablation Study of Architectural Components (Mean $\pm$ SD). Boxed cells (\fbox{...}) indicate selected configurations.}
\label{tab:ablation_cases}
\centering
\scriptsize
\setlength{\tabcolsep}{1.5pt}
\setlength{\fboxsep}{1.0pt}
\renewcommand{\arraystretch}{0.95}
\begin{tabular}{l l *{4}{c} | l *{4}{c} | l *{4}{c}}
\toprule
 & \multicolumn{5}{c}{\textbf{BATADAL}} & \multicolumn{5}{c}{\textbf{SWAT}} & \multicolumn{5}{c}{\textbf{WADI}} \\
\cmidrule(lr){2-6}\cmidrule(lr){7-11}\cmidrule(lr){12-16}
\textbf{Method} & \textbf{Params} & \textbf{AC@1} & \textbf{AC@5} & \textbf{AC@10} & \textbf{auc@10} & \textbf{Params} & \textbf{AC@1} & \textbf{AC@5} & \textbf{AC@10} & \textbf{auc@10} & \textbf{Params} & \textbf{AC@1} & \textbf{AC@5} & \textbf{AC@10} & \textbf{auc@10} \\
\midrule
\multicolumn{10}{l}{\textit{Loss Ablations}} \\
\midrule
w/o Sparse & \textit{279.4K} & \cellcolor{yellow!0!red!35}0.381{\tiny$\pm$0.08} & \cellcolor{green!100!yellow!35}0.181{\tiny$\pm$0.02} & \cellcolor{green!100!yellow!35}0.247{\tiny$\pm$0.04} & \cellcolor{green!100!yellow!35}0.238{\tiny$\pm$0.03} & \textit{74.2K} & \cellcolor{green!75!yellow!35}0.217{\tiny$\pm$0.03} & \cellcolor{yellow!51!red!35}0.333{\tiny$\pm$0.03} & \cellcolor{yellow!20!red!35}0.400{\tiny$\pm$0.00} & \cellcolor{yellow!50!red!35}0.338{\tiny$\pm$0.02} & \textit{84.5K} & \cellcolor{yellow!0!red!35}0.133{\tiny$\pm$0.00} & \cellcolor{yellow!88!red!35}0.278{\tiny$\pm$0.02} & \cellcolor{yellow!46!red!35}0.319{\tiny$\pm$0.01} & \cellcolor{yellow!40!red!35}0.265{\tiny$\pm$0.01} \\
w/o Smooth & \textit{279.4K} & \cellcolor{yellow!0!red!35}0.381{\tiny$\pm$0.08} & \cellcolor{green!100!yellow!35}0.181{\tiny$\pm$0.02} & \cellcolor{green!100!yellow!35}0.247{\tiny$\pm$0.04} & \cellcolor{green!100!yellow!35}0.238{\tiny$\pm$0.03} & \textit{74.2K} & \cellcolor{yellow!99!red!35}0.167{\tiny$\pm$0.03} & \cellcolor{yellow!96!red!35}0.383{\tiny$\pm$0.03} & \cellcolor{yellow!60!red!35}0.433{\tiny$\pm$0.06} & \cellcolor{yellow!74!red!35}0.353{\tiny$\pm$0.01} & \textit{84.5K} & \cellcolor{yellow!0!red!35}0.133{\tiny$\pm$0.00} & \cellcolor{green!44!yellow!35}0.289{\tiny$\pm$0.02} & \cellcolor{green!100!yellow!35}0.404{\tiny$\pm$0.03} & \cellcolor{green!44!yellow!35}0.294{\tiny$\pm$0.02} \\
w/o KL & \textit{279.4K} & \cellcolor{green!100!yellow!35}0.476{\tiny$\pm$0.08} & \cellcolor{yellow!100!red!35}0.162{\tiny$\pm$0.02} & \cellcolor{yellow!85!red!35}0.213{\tiny$\pm$0.01} & \cellcolor{green!73!yellow!35}0.234{\tiny$\pm$0.01} & \textit{74.2K} & \cellcolor{green!75!yellow!35}0.217{\tiny$\pm$0.03} & \cellcolor{green!11!yellow!35}0.400{\tiny$\pm$0.05} & \cellcolor{yellow!60!red!35}0.433{\tiny$\pm$0.03} & \cellcolor{yellow!96!red!35}0.367{\tiny$\pm$0.04} & \textit{84.5K} & \cellcolor{yellow!0!red!35}0.133{\tiny$\pm$0.00} & \cellcolor{yellow!55!red!35}0.271{\tiny$\pm$0.01} & \cellcolor{yellow!0!red!35}0.293{\tiny$\pm$0.04} & \cellcolor{yellow!43!red!35}0.266{\tiny$\pm$0.02} \\
\midrule
\multicolumn{10}{l}{\textit{Architecture Ablations}} \\
\midrule
w/o Orth & \textit{279.4K} & \cellcolor{green!0!yellow!35}0.429{\tiny$\pm$0.00} & \cellcolor{yellow!0!red!35}0.143{\tiny$\pm$0.03} & \cellcolor{yellow!72!red!35}0.209{\tiny$\pm$0.02} & \cellcolor{yellow!0!red!35}0.210{\tiny$\pm$0.02} & \textit{74.2K} & \cellcolor{green!75!yellow!35}0.217{\tiny$\pm$0.03} & \cellcolor{yellow!51!red!35}0.333{\tiny$\pm$0.06} & \cellcolor{yellow!0!red!35}0.383{\tiny$\pm$0.03} & \cellcolor{yellow!23!red!35}0.322{\tiny$\pm$0.03} & \textit{84.5K} & \cellcolor{yellow!0!red!35}0.133{\tiny$\pm$0.00} & \cellcolor{yellow!88!red!35}0.278{\tiny$\pm$0.02} & \cellcolor{yellow!20!red!35}0.304{\tiny$\pm$0.01} & \cellcolor{yellow!36!red!35}0.264{\tiny$\pm$0.01} \\
w/o Mixer & \textit{278.4K} & \cellcolor{green!0!yellow!35}0.429{\tiny$\pm$0.00} & \cellcolor{yellow!100!red!35}0.162{\tiny$\pm$0.02} & \cellcolor{yellow!85!red!35}0.213{\tiny$\pm$0.01} & \cellcolor{green!21!yellow!35}0.227{\tiny$\pm$0.01} & \textit{73.7K} & \cellcolor{yellow!49!red!35}0.133{\tiny$\pm$0.03} & \cellcolor{green!11!yellow!35}0.400{\tiny$\pm$0.05} & \cellcolor{green!20!yellow!35}0.483{\tiny$\pm$0.06} & \cellcolor{yellow!93!red!35}0.365{\tiny$\pm$0.03} & \textit{84.0K} & \cellcolor{yellow!0!red!35}0.133{\tiny$\pm$0.00} & \cellcolor{green!100!yellow!35}0.300{\tiny$\pm$0.03} & \cellcolor{green!73!yellow!35}0.389{\tiny$\pm$0.05} & \cellcolor{green!44!yellow!35}0.294{\tiny$\pm$0.02} \\
w/o L-Attn & \textit{82.8K} & \cellcolor{green!0!yellow!35}0.429{\tiny$\pm$0.00} & \cellcolor{green!100!yellow!35}0.181{\tiny$\pm$0.02} & \cellcolor{green!18!yellow!35}0.223{\tiny$\pm$0.02} & \cellcolor{green!82!yellow!35}0.236{\tiny$\pm$0.02} & \textit{25.1K} & \cellcolor{yellow!0!red!35}0.100{\tiny$\pm$0.00} & \cellcolor{green!100!yellow!35}0.500{\tiny$\pm$0.00} & \cellcolor{green!100!yellow!35}0.550{\tiny$\pm$0.00} & \cellcolor{green!100!yellow!35}0.430{\tiny$\pm$0.01} & \textit{35.3K} & \cellcolor{green!100!yellow!35}\fbox{0.333{\tiny$\pm$0.00}} & \cellcolor{green!44!yellow!35}\fbox{0.289{\tiny$\pm$0.02}} & \cellcolor{green!53!yellow!35}\fbox{0.378{\tiny$\pm$0.02}} & \cellcolor{green!100!yellow!35}\fbox{0.310{\tiny$\pm$0.01}} \\
Pred (MLP) & \textit{414.9K} & \cellcolor{green!0!yellow!35}0.429{\tiny$\pm$0.00} & \cellcolor{green!100!yellow!35}0.181{\tiny$\pm$0.02} & \cellcolor{yellow!0!red!35}0.188{\tiny$\pm$0.04} & \cellcolor{green!3!yellow!35}0.225{\tiny$\pm$0.02} & \textit{108.1K} & \cellcolor{green!100!yellow!35}\fbox{0.233{\tiny$\pm$0.03}} & \cellcolor{yellow!81!red!35}\fbox{0.367{\tiny$\pm$0.08}} & \cellcolor{yellow!40!red!35}\fbox{0.417{\tiny$\pm$0.06}} & \cellcolor{yellow!63!red!35}\fbox{0.347{\tiny$\pm$0.04}} & \textit{118.4K} & \cellcolor{yellow!0!red!35}0.133{\tiny$\pm$0.00} & \cellcolor{yellow!33!red!35}0.267{\tiny$\pm$0.03} & \cellcolor{yellow!39!red!35}0.315{\tiny$\pm$0.05} & \cellcolor{yellow!25!red!35}0.261{\tiny$\pm$0.02} \\
Pred (Linear) & \textit{279.4K} & \cellcolor{green!100!yellow!35}\fbox{0.476{\tiny$\pm$0.08}} & \cellcolor{yellow!100!red!35}\fbox{0.162{\tiny$\pm$0.02}} & \cellcolor{yellow!85!red!35}\fbox{0.213{\tiny$\pm$0.01}} & \cellcolor{green!73!yellow!35}\fbox{0.234{\tiny$\pm$0.01}} & \textit{74.2K} & \cellcolor{green!75!yellow!35}0.217{\tiny$\pm$0.06} & \cellcolor{yellow!51!red!35}0.333{\tiny$\pm$0.03} & \cellcolor{yellow!40!red!35}0.417{\tiny$\pm$0.03} & \cellcolor{yellow!23!red!35}0.322{\tiny$\pm$0.01} & \textit{84.5K} & \cellcolor{yellow!0!red!35}0.133{\tiny$\pm$0.00} & \cellcolor{yellow!0!red!35}0.260{\tiny$\pm$0.01} & \cellcolor{yellow!13!red!35}0.300{\tiny$\pm$0.02} & \cellcolor{yellow!0!red!35}0.254{\tiny$\pm$0.01} \\
\bottomrule
\end{tabular}
\end{table*}

We evaluate the contribution of the major loss and architectural components of SFlexRCA through ablation studies on BATADAL, SWaT, and WADI. To isolate the effect of the loss formulation, all loss ablations use a common reference architecture with the linear prediction head and linear attention enabled. Starting from this fixed architecture, we remove one loss component at a time while keeping all other architectural and training settings unchanged. Architectural ablations are conducted separately to examine the contribution of the orthogonal representation, temporal mixer, linear attention, and prediction head and to determine the dataset-specific configurations reported in Table~IV. Thus, the parameter counts in the loss-ablation results correspond to the fixed reference architecture rather than necessarily to the final selected configuration for each dataset.

\textbf{Contribution of Loss Components.}
The loss ablations show that the sparse and smoothness losses are particularly important for the more complex SWaT and WADI datasets. Removing the sparse loss reduces AC@1 from 0.233 to 0.217 on SWaT and from 0.333 to 0.133 on WADI, indicating that sparsity helps isolate the variables most strongly associated with failures. Similarly, removing the smoothness loss reduces AC@1 to 0.167 on SWaT and 0.133 on WADI, suggesting that temporally stable interaction estimates become more important as temporal dependencies become more pronounced. The KL loss has its strongest effect on WADI, where removing it reduces AC@10 from 0.378 to 0.293. Its effect on BATADAL is limited: removing the KL term preserves AC@1 at 0.476 and produces comparable overall performance, consistent with the relatively simpler dependency structure of BATADAL, therefore in our experiments we maintain the KL loss for WADI and SWaT but remove it for BATADAL. Overall, the results indicate that the loss components are complementary, with their importance depending on the temporal and structural complexity of the dataset.

\textbf{Contribution of Architectural Components.}
The architectural ablations highlight the complementary roles of orthogonal representations, temporal mixing, linear attention, and the prediction head. Removing the orthogonal coefficients consistently degrades performance across the three datasets, particularly on BATADAL and WADI, supporting their importance for constructing informative variable interactions. The temporal mixer primarily benefits top-ranked localization: removing it reduces AC@1 from 0.233 to 0.133 on SWAT and from 0.333 to 0.133 on WADI, indicating that temporal mixing provides useful information for distinguishing variables associated with failures.
The contribution of linear attention is dataset-dependent. While BATADAL and SWAT retain linear attention in their selected configurations, WADI achieves its best overall RCA performance without it, demonstrating that SFlexRCA can achieve effective localization using its core orthogonal representation and lightweight prediction components alone. This architectural flexibility also supports scalability across systems with different monitoring dimensionalities: although WADI has the largest number of monitored sensors ($P=127$), its selected configuration requires only 35.3K parameters. Rather than imposing increasing architectural complexity as the number of sensors grows, SFlexRCA can adapt its components to the characteristics of each system, highlighting its flexibility and scalability across heterogeneous IIoT environments.
The prediction head also exhibits dataset-dependent behavior. The linear prediction head is used as the default configuration because it provides a more compact model, while the MLP prediction head is considered as an alternative in the ablation study. Accordingly, Pred (Linear) denotes the default linear prediction head with linear attention retained, whereas Pred (MLP) replaces the prediction head with an MLP while keeping linear attention enabled. The w/o L-Attn configuration, in contrast, removes the linear attention module while retaining the default linear prediction head. Based on the results, the linear prediction head is used for BATADAL and WADI, whereas the MLP prediction head provides the selected configuration for SWAT. Overall, these results highlight the flexibility of SFlexRCA to adapt its architecture to dataset-specific characteristics while balancing RCA accuracy and model complexity.

\textbf{Summary.} Overall, the ablation results demonstrate that SFlexRCA provides both **architectural flexibility and scalability**. Its components can be selectively enabled or simplified according to dataset characteristics, allowing RCA accuracy to be balanced against model complexity. This adaptability enables SFlexRCA to remain compact across systems with different monitoring dimensionalities, making it well suited to resource-constrained edge environments.

\subsection{RQ-4: Sensitivity to Temporal Context}

\begin{figure*}[t]
    \centering

    \hfill
    \begin{subfigure}[b]{0.32\textwidth}
    	\centering
    	\includegraphics[width=\linewidth]{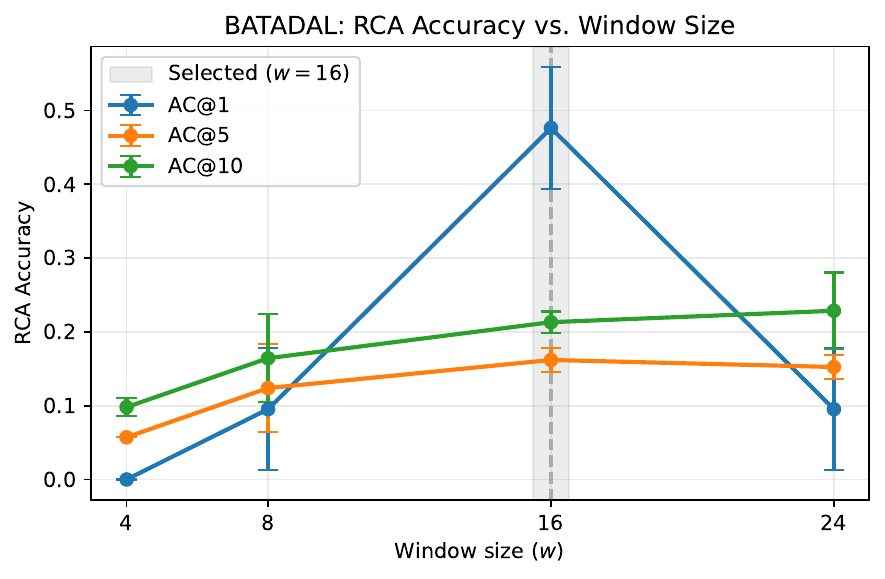}
    	\caption{BATADAL: RCA vs. window size.}
    	\label{fig:rq4_batadal_window_size}
    \end{subfigure}
    \hfill
    \begin{subfigure}[b]{0.32\textwidth}
        \centering
        \includegraphics[width=\linewidth]{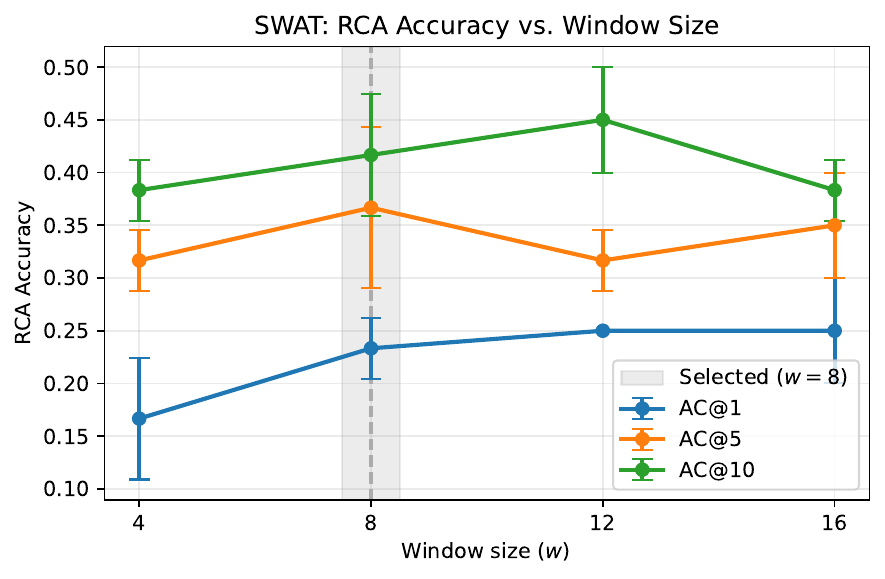}
        \caption{SWaT: RCA vs. window size.}
        \label{fig:rq4_swat_window_size}
    \end{subfigure}
    \hfill
    \begin{subfigure}[b]{0.32\textwidth}
        \centering
        \includegraphics[width=\linewidth]{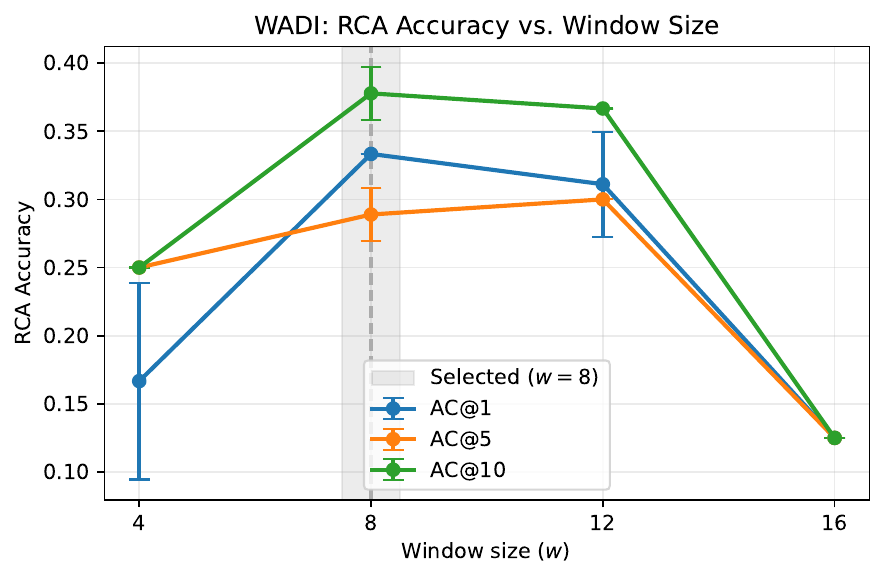}
        \caption{WADI: RCA vs. window size.}
        \label{fig:rq4_wadi_window_size}
    \end{subfigure}
    \hfill
    \vspace{0.8em}

    \caption{
        Sensitivity analysis of RCA accuracy with respect to window size for BATADAL, SWaT and WADI datasets.
    }
    \label{fig:rq2_efficiency}
\end{figure*}
We further examine the sensitivity of SFlexRCA to temporal window size to determine how much historical context is required for effective RCA. The results show that the optimal temporal context is dataset-dependent. We evaluate windows of 4, 8, 12, and 16 for SWaT and WADI (4--16 seconds), and 4, 8, 16, and 24 for BATADAL (4--24 hours), reflecting their different sampling frequencies as summarized in Table~\ref{tab:dataset_comparison}.

\textbf{Effect of Temporal Context.}
On BATADAL, increasing the window size from 4 to 16 improves AC@1 from 0.0 to 0.4762, the highest value observed. Although AC@1 decreases at larger windows, AC@5 remains stable from window 8 onward and AC@10 continues improving through window 24. We therefore select a window size of 16 based on its superior top-ranked localization performance. SWaT exhibits a similar benefit from additional context, with AC@1 increasing from 0.167 at a window of 4 to 0.233 at 8 and reaching 0.250 at 12 and 16. However, the window of 8 already provides a strong performance level while requiring less historical context; we therefore select the earliest window achieving competitive performance to reduce the amount of context required for RCA. WADI exhibits a different behavior: performance improves from a window of 4 to 8, reaching AC@1 of 0.333 and AC@10 of 0.378, but decreases substantially at a window of 16, where all three metrics fall to 0.125. This suggests that excessive temporal context can introduce less relevant historical information and make it more difficult to isolate the measurements associated with the current failure. Therefore, we select a window of 8 for WADI to balance the benefits of temporal context with the risk of including less relevant historical information.

\textbf{Summary.}
Overall, the sensitivity analysis demonstrates that SFlexRCA benefits from dataset-specific temporal context rather than simply favoring larger windows. BATADAL benefits from a longer intermediate context, while SWaT achieves strong performance with an earlier window size and WADI favors a shorter context. This supports the flexibility of SFlexRCA to adapt its temporal receptive field to the dynamics of the monitored system while avoiding unnecessary historical context.

\subsection{RQ-5: Effect of Dataset-Specific Orthogonal Projections}

\begin{table*}[t]
\caption{Root Cause Analysis Accuracy Across Different Projections For BATADAL and SWAT and WADI (Mean $\pm$ SD).}
\label{tab:RQ5}
\centering
\scriptsize
\setlength{\tabcolsep}{2.0pt}
\renewcommand{\arraystretch}{0.92}
\setlength{\fboxsep}{1.0pt}
\begin{tabular}{l *{4}{c} !{\vrule width 1.2pt} *{4}{c} !{\vrule width 1.2pt} *{4}{c}}
\toprule
 & \multicolumn{4}{c}{\textbf{BATADAL}  } & \multicolumn{4}{c}{\textbf{SWAT} } & \multicolumn{4}{c}{\textbf{WADI} } \\
\cmidrule(lr){2-5}\cmidrule(lr){6-9}\cmidrule(lr){10-13}
\textbf{Method} & \textbf{AC@1} & \textbf{AC@5} & \textbf{AC@10} & \textbf{auc@10} & \textbf{AC@1} & \textbf{AC@5} & \textbf{AC@10} & \textbf{auc@10} & \textbf{AC@1} & \textbf{AC@5} & \textbf{AC@10} & \textbf{auc@10} \\
\midrule
\multicolumn{1}{l}{\textbf{Baselines}} \\
none & \cellcolor{green!33!yellow!35}0.429{\tiny$\pm$0.00} & \cellcolor{yellow!0!red!35}0.143{\tiny$\pm$0.03} & \cellcolor{green!77!yellow!35}0.209{\tiny$\pm$0.02} & \cellcolor{yellow!85!red!35}0.210{\tiny$\pm$0.02} & \cellcolor{green!50!yellow!35}0.217{\tiny$\pm$0.06} & \cellcolor{yellow!49!red!35}0.317{\tiny$\pm$0.03} & \cellcolor{yellow!0!red!35}0.400{\tiny$\pm$0.00} & \cellcolor{yellow!17!red!35}0.308{\tiny$\pm$0.01} & \cellcolor{green!0!yellow!35}0.333{\tiny$\pm$0.00} & \cellcolor{yellow!0!red!35}0.289{\tiny$\pm$0.02} & \cellcolor{yellow!0!red!35}0.367{\tiny$\pm$0.00} & \cellcolor{yellow!72!red!35}0.309{\tiny$\pm$0.00} \\
learned & \cellcolor{yellow!66!red!35}0.381{\tiny$\pm$0.08} & \cellcolor{yellow!33!red!35}0.149{\tiny$\pm$0.03} & \cellcolor{yellow!0!red!35}0.177{\tiny$\pm$0.05} & \cellcolor{yellow!6!red!35}0.193{\tiny$\pm$0.05} & \cellcolor{green!50!yellow!35}0.217{\tiny$\pm$0.03} & \cellcolor{green!100!yellow!35}0.367{\tiny$\pm$0.03} & \cellcolor{yellow!66!red!35}0.417{\tiny$\pm$0.03} & \cellcolor{green!100!yellow!35}0.360{\tiny$\pm$0.02} & \cellcolor{green!0!yellow!35}0.333{\tiny$\pm$0.00} & \cellcolor{yellow!0!red!35}0.289{\tiny$\pm$0.02} & \cellcolor{yellow!0!red!35}0.367{\tiny$\pm$0.00} & \cellcolor{yellow!54!red!35}0.308{\tiny$\pm$0.00} \\
\midrule
\multicolumn{1}{l}{\textbf{Static Projections}} \\
legendre & \cellcolor{yellow!66!red!35}0.381{\tiny$\pm$0.08} & \cellcolor{yellow!49!red!35}0.152{\tiny$\pm$0.04} & \cellcolor{yellow!4!red!35}0.178{\tiny$\pm$0.03} & \cellcolor{yellow!56!red!35}0.204{\tiny$\pm$0.03} & \cellcolor{green!0!yellow!35}0.200{\tiny$\pm$0.00} & \cellcolor{yellow!99!red!35}0.333{\tiny$\pm$0.03} & \cellcolor{yellow!0!red!35}0.400{\tiny$\pm$0.05} & \cellcolor{yellow!52!red!35}0.318{\tiny$\pm$0.01} & \cellcolor{green!0!yellow!35}0.333{\tiny$\pm$0.00} & \cellcolor{yellow!0!red!35}0.289{\tiny$\pm$0.02} & \cellcolor{yellow!0!red!35}0.367{\tiny$\pm$0.00} & \cellcolor{yellow!36!red!35}0.307{\tiny$\pm$0.01} \\
laguerre & \cellcolor{yellow!66!red!35}0.381{\tiny$\pm$0.08} & \cellcolor{yellow!99!red!35}0.162{\tiny$\pm$0.03} & \cellcolor{yellow!4!red!35}0.178{\tiny$\pm$0.03} & \cellcolor{yellow!78!red!35}0.208{\tiny$\pm$0.03} & \cellcolor{green!0!yellow!35}0.200{\tiny$\pm$0.00} & \cellcolor{yellow!0!red!35}0.300{\tiny$\pm$0.05} & \cellcolor{yellow!0!red!35}0.400{\tiny$\pm$0.00} & \cellcolor{yellow!0!red!35}0.303{\tiny$\pm$0.02} & \cellcolor{green!0!yellow!35}0.333{\tiny$\pm$0.00} & \cellcolor{yellow!0!red!35}0.289{\tiny$\pm$0.02} & \cellcolor{yellow!0!red!35}0.367{\tiny$\pm$0.00} & \cellcolor{yellow!0!red!35}0.304{\tiny$\pm$0.01} \\
chebyshev & \cellcolor{yellow!0!red!35}0.333{\tiny$\pm$0.08} & \cellcolor{yellow!0!red!35}0.143{\tiny$\pm$0.03} & \cellcolor{yellow!73!red!35}0.190{\tiny$\pm$0.05} & \cellcolor{yellow!0!red!35}0.192{\tiny$\pm$0.06} & \cellcolor{green!100!yellow!35}0.233{\tiny$\pm$0.03} & \cellcolor{yellow!49!red!35}0.317{\tiny$\pm$0.03} & \cellcolor{green!33!yellow!35}0.433{\tiny$\pm$0.03} & \cellcolor{green!17!yellow!35}0.337{\tiny$\pm$0.01} & \cellcolor{green!0!yellow!35}0.333{\tiny$\pm$0.00} & \cellcolor{green!100!yellow!35}0.300{\tiny$\pm$0.00} & \cellcolor{yellow!0!red!35}0.367{\tiny$\pm$0.00} & \cellcolor{yellow!90!red!35}0.310{\tiny$\pm$0.01} \\
hermite & \cellcolor{yellow!66!red!35}0.381{\tiny$\pm$0.08} & \cellcolor{yellow!49!red!35}0.152{\tiny$\pm$0.04} & \cellcolor{yellow!4!red!35}0.178{\tiny$\pm$0.03} & \cellcolor{yellow!56!red!35}0.204{\tiny$\pm$0.03} & \cellcolor{yellow!0!red!35}0.167{\tiny$\pm$0.03} & \cellcolor{yellow!99!red!35}0.333{\tiny$\pm$0.08} & \cellcolor{yellow!0!red!35}0.400{\tiny$\pm$0.05} & \cellcolor{yellow!47!red!35}0.317{\tiny$\pm$0.03} & \cellcolor{green!0!yellow!35}0.333{\tiny$\pm$0.00} & \cellcolor{green!100!yellow!35}0.300{\tiny$\pm$0.00} & \cellcolor{yellow!0!red!35}0.367{\tiny$\pm$0.00} & \cellcolor{green!100!yellow!35}0.317{\tiny$\pm$0.00} \\
fourier & \cellcolor{green!33!yellow!35}0.429{\tiny$\pm$0.00} & \cellcolor{green!100!yellow!35}0.181{\tiny$\pm$0.04} & \cellcolor{green!26!yellow!35}0.200{\tiny$\pm$0.01} & \cellcolor{green!47!yellow!35}0.223{\tiny$\pm$0.02} & \cellcolor{green!50!yellow!35}0.217{\tiny$\pm$0.03} & \cellcolor{yellow!99!red!35}0.333{\tiny$\pm$0.03} & \cellcolor{green!100!yellow!35}0.450{\tiny$\pm$0.05} & \cellcolor{green!35!yellow!35}0.342{\tiny$\pm$0.02} & \cellcolor{green!0!yellow!35}0.333{\tiny$\pm$0.00} & \cellcolor{yellow!0!red!35}0.289{\tiny$\pm$0.02} & \cellcolor{green!100!yellow!35}0.389{\tiny$\pm$0.04} & \cellcolor{green!9!yellow!35}0.311{\tiny$\pm$0.01} \\
\midrule
\multicolumn{1}{l}{\textbf{Dataset Specific Projection}} \\
orthogonal (proposed) & \cellcolor{green!100!yellow!35}0.476{\tiny$\pm$0.08} & \cellcolor{green!0!yellow!35}0.162{\tiny$\pm$0.02} & \cellcolor{green!100!yellow!35}0.213{\tiny$\pm$0.01} & \cellcolor{green!100!yellow!35}0.234{\tiny$\pm$0.01} & \cellcolor{green!100!yellow!35}0.233{\tiny$\pm$0.03} & \cellcolor{green!100!yellow!35}0.367{\tiny$\pm$0.08} & \cellcolor{yellow!66!red!35}0.417{\tiny$\pm$0.06} & \cellcolor{green!52!yellow!35}0.347{\tiny$\pm$0.04} & \cellcolor{green!0!yellow!35}0.333{\tiny$\pm$0.00} & \cellcolor{yellow!0!red!35}0.289{\tiny$\pm$0.02} & \cellcolor{green!0!yellow!35}0.378{\tiny$\pm$0.02} & \cellcolor{yellow!90!red!35}0.310{\tiny$\pm$0.01} \\
\bottomrule
\end{tabular}
\end{table*}

\begin{figure*}[t]
    \centering
    \begin{subfigure}[b]{0.24\textwidth}
        \centering
        \includegraphics[width=\textwidth]{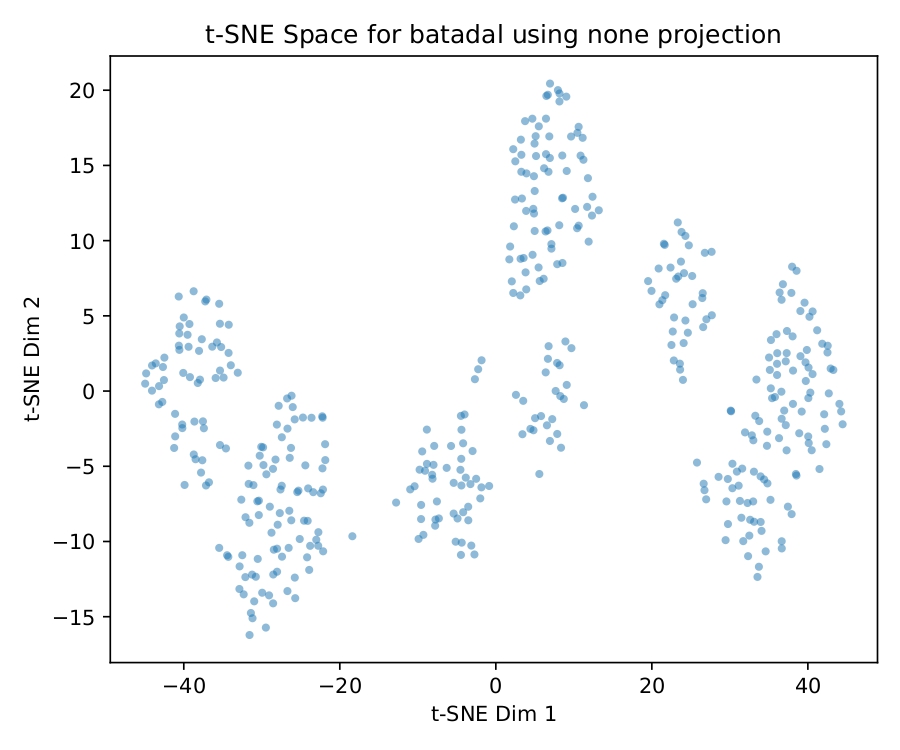}
        \caption{None (Baseline)}
        \label{fig:tsne_none}
    \end{subfigure}
    \hfill
    \begin{subfigure}[b]{0.24\textwidth}
        \centering
        \includegraphics[width=\textwidth]{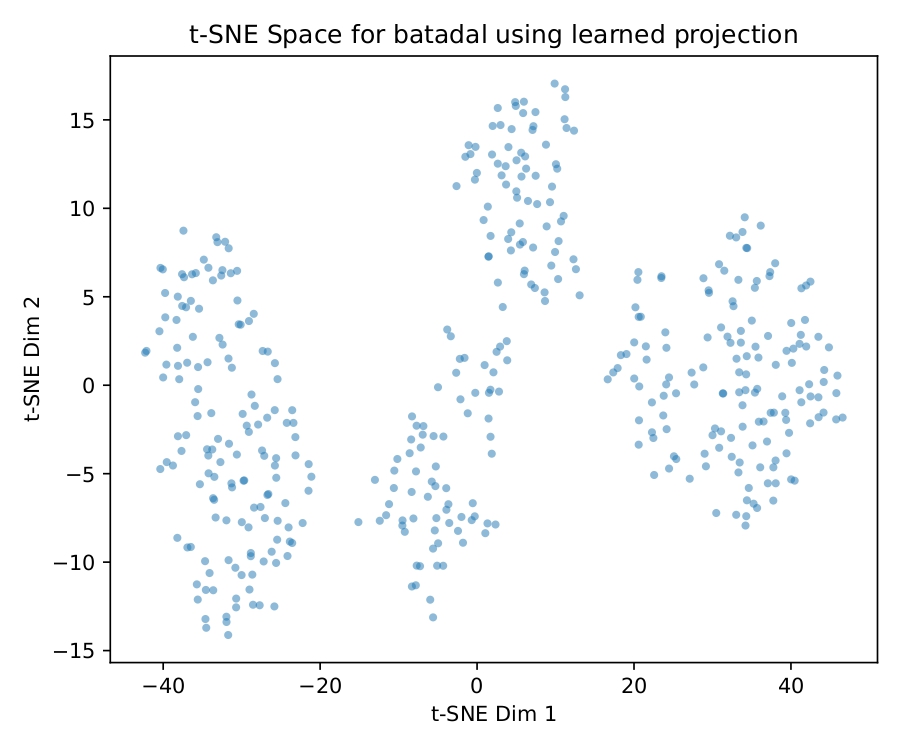}
        \caption{Learned}
        \label{fig:tsne_learned}
    \end{subfigure}
    \hfill
    \begin{subfigure}[b]{0.24\textwidth}
        \centering
        \includegraphics[width=\textwidth]{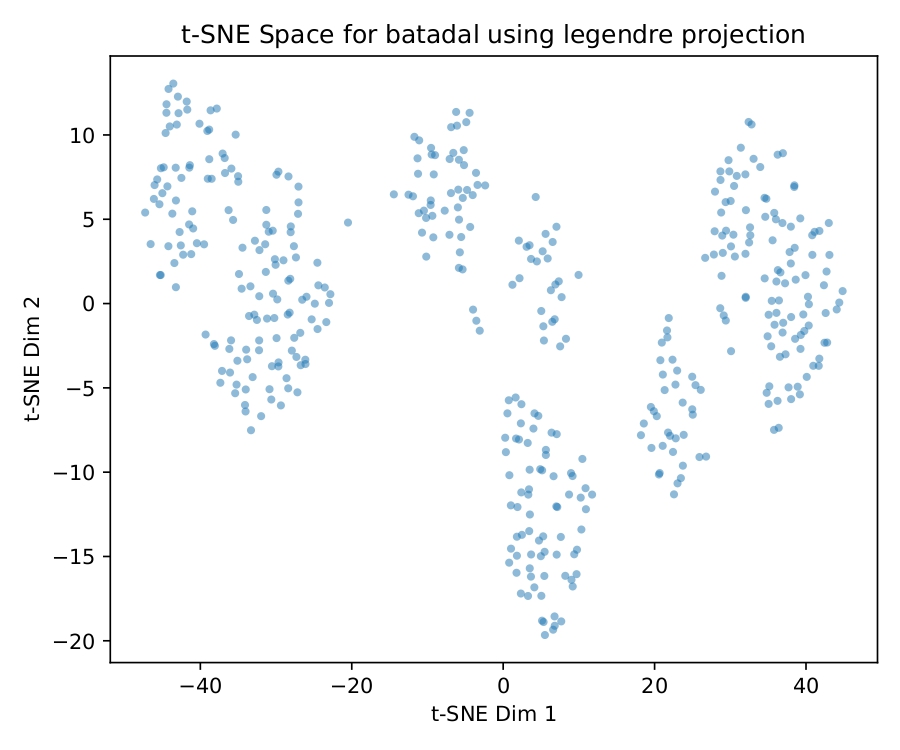}
        \caption{Legendre}
        \label{fig:tsne_legendre}
    \end{subfigure}
    \hfill
    \begin{subfigure}[b]{0.24\textwidth}
        \centering
        \includegraphics[width=\textwidth]{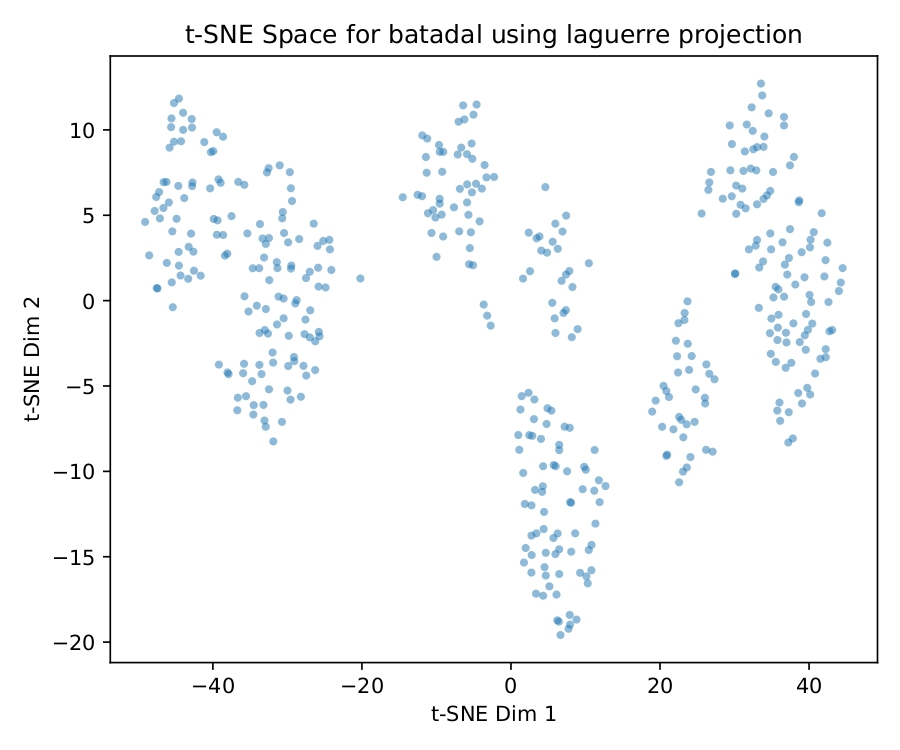}
        \caption{Laguerre}
        \label{fig:tsne_laguerre}
    \end{subfigure}

    \vspace{0.8em} 

    \begin{subfigure}[b]{0.24\textwidth}
        \centering
        \includegraphics[width=\textwidth]{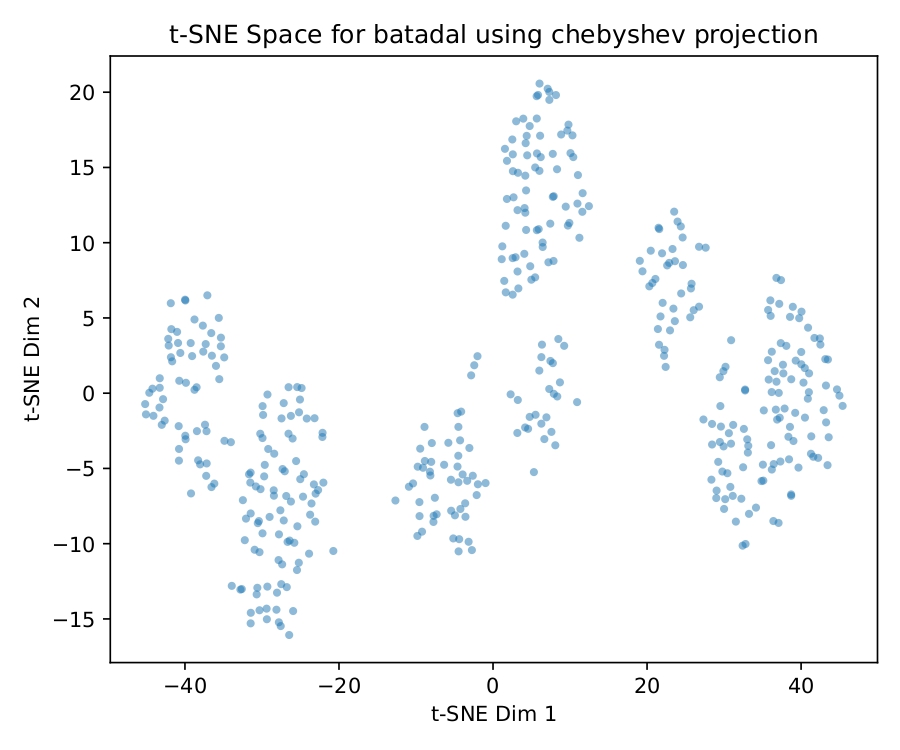}
        \caption{Chebyshev}
        \label{fig:tsne_chebyshev}
    \end{subfigure}
    \hfill
    \begin{subfigure}[b]{0.24\textwidth}
        \centering
        \includegraphics[width=\textwidth]{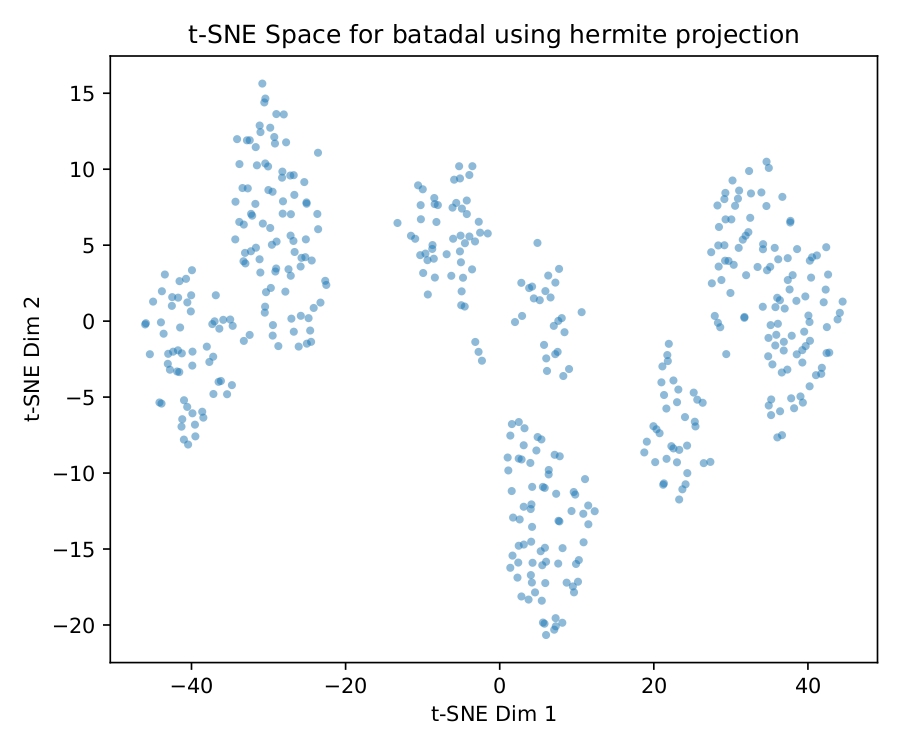}
        \caption{Hermite}
        \label{fig:tsne_hermite}
    \end{subfigure}
    \hfill
    \begin{subfigure}[b]{0.24\textwidth}
        \centering
        \includegraphics[width=\textwidth]{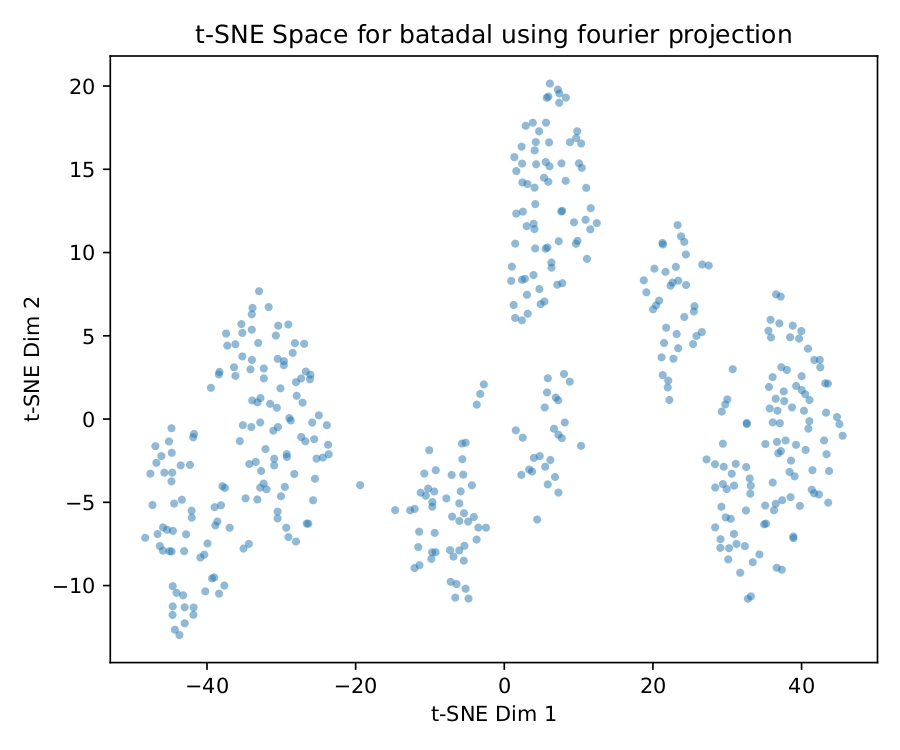}
        \caption{Fourier}
        \label{fig:tsne_fourier}
    \end{subfigure}
    \hfill
    \begin{subfigure}[b]{0.24\textwidth}
        \centering
        \includegraphics[width=\textwidth]{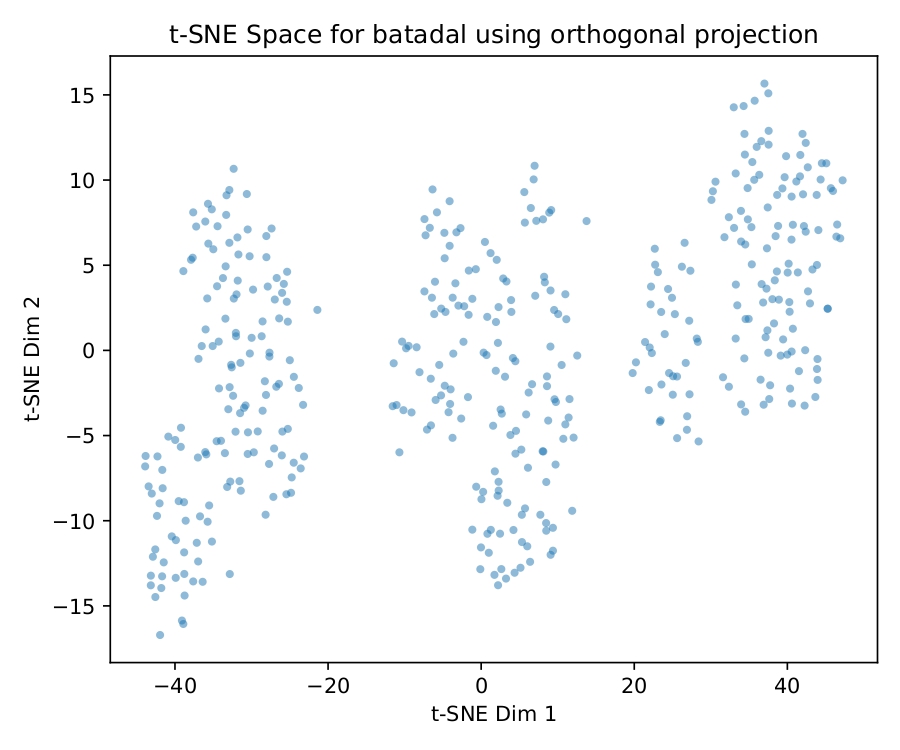}
        \caption{Orthogonal (Proposed)}
        \label{fig:tsne_orthogonal}
    \end{subfigure}

    \caption{t-SNE latent feature projections on BATADAL corresponding to baseline methods, static polynomial/frequency projections, and our proposed orthogonal projection (matching Table~\ref{tab:RQ5}).}
    \label{fig:tsne_grid}
\end{figure*}

We evaluate the effect of the proposed dataset-specific orthogonal projection by replacing it with alternative representations while keeping the remaining SFlexRCA architecture unchanged. Specifically, we compare the proposed projection against no projection (i.e., none), a learned projection, and several fixed basis transformations, including Legendre, Laguerre, Chebyshev, Hermite, and Fourier representations following the experimental methodology of~\cite{280-FreqTimeLoss,olinear}.

\textbf{Comparison of Projection Methods:}
Table~\ref{tab:RQ5} shows that the dataset-specific orthogonal projection provides the strongest overall RCA performance. On BATADAL, the proposed projection achieves the highest AC@1, AC@10, and AUC@10, reaching $0.476$, $0.213$, and $0.234$, respectively. On SWaT, it ties for the highest AC@1 and AC@5. The gains are more modest on WADI, where the different projections produce similar results, but the proposed projection remains competitive across all metrics.
The observed differences are consistent with how the projection is constructed. Unlike fixed bases, the proposed transformation is derived from the temporal covariance structure of normal operating data for each dataset. Consequently, the resulting orthogonal coordinates are adapted to the characteristic dynamics of the monitored IIoT system rather than imposing a universal temporal basis. This distinction is particularly relevant because the datasets differ substantially in sampling frequency and temporal characteristics (Table~\ref{tab:dataset_comparison}). BATADAL is sampled hourly, whereas SWaT and WADI are sampled at one-second intervals; therefore, the temporal correlations captured by the learned basis can differ substantially across systems.
The results also provide evidence for the importance of orthogonality in addition to dataset adaptation. The learned projection is dataset-adaptive but does not impose an orthogonal basis, whereas the proposed projection combines dataset adaptation with orthogonality. The proposed method generally outperforms the learned projection on BATADAL and SWaT, while achieving competitive or superior performance relative to the fixed orthogonal bases. This suggests that the observed benefit is not solely due to adapting the projection to the dataset, but also to organizing the adapted representation into orthogonal components.
This property is particularly relevant to IIoT telemetry, where sensor measurements can exhibit substantial redundancy. The orthogonal basis reduces such redundancy while preserving the information captured by the normal-operation covariance structure, providing a more separated representation for subsequent latent modeling and residual-based localization.

\textbf{TSNE Visualization of Projections:}
To visually inspect how different transformations structure the feature space, Figure~\ref{fig:tsne_grid} presents the t-SNE manifold projections of the latent representations on BATADAL across baseline, static polynomial/frequency, and proposed orthogonal projections following the experimental methodology of~\cite{vkd_367}. 
As shown in Figures~\ref{fig:tsne_none} and \ref{fig:tsne_learned}, both the unprojected baseline (\textit{none}) and the \textit{learned} non-orthogonal projection produce fragmented and dispersed feature clusters. This structural dispersion mirrors their lower localization accuracy in Table~\ref{tab:RQ5}, where non-orthogonal learned features fail to maintain distinct boundaries between normal operational modes and anomalous deviations. 
In contrast, static orthogonal polynomial bases (Legendre, Laguerre, Chebyshev, Hermite) and frequency transformations (Fourier) shown in Figures~\ref{fig:tsne_legendre}–\ref{fig:tsne_fourier} enforce a more structured latent topology. However, because these bases impose fixed, system-agnostic assumptions over time, the resulting clusters remain partially overlapping or stretched across non-linearly correlated dimensions. 
Our proposed dataset-specific orthogonal projection (Figure~\ref{fig:tsne_orthogonal}) achieves the highest degree of cluster compactness and inter-cluster separation. By constructing an orthogonal basis directly from the system's empirical covariance structure, it effectively decorrelates redundant sensor channels while preserving global system dynamics. This visual compactness directly corroborates the superior top-$k$ accuracy (AC@1 = $0.476$, AUC@10 = $0.234$) observed on BATADAL, confirming that joint dataset adaptation and strict orthogonality yield cleaner decision spaces for residual localization.

\textbf{Summary:}
In summary, the empirical metrics and qualitative t-SNE visualizations jointly validate the core design of our proposed representation. While dataset-specific adaptation tailors the transformation to the unique sampling frequencies and temporal dynamics of individual IIoT deployments, orthogonality enforces strict feature decorrelation across redundant sensor channels. Geometrically, this dual constraint eliminates cluster dispersion and produces tight, well-separated latent representations—preventing failure modes from bleeding into normal operating regions. For IIoT systems characterized by high-dimensional telemetry and inter-variable coupling, this structural preservation is essential: it ensures that subtle anomaly residuals remain localized and distinct, thereby delivering superior top-$k$ diagnostic precision over both static, system-agnostic bases and unconstrained learned projections.

\section{Threats to Validity}

\textbf{Internal Validity:}
Potential threats arise from implementation choices, hyperparameter tuning, and differences in the optimization requirements of competing RCA methods. We mitigate these effects through a consistent evaluation pipeline, multiple random seeds, and ablation and sensitivity analyses.

\textbf{External Validity:}
We evaluate SFlexRCA on BATADAL, SWAT, and WADI, which cover different telemetry dimensionalities, temporal characteristics, and training-data regimes. However, these benchmarks may not capture the full diversity of real IIoT deployments, including noisier telemetry, incomplete fault labels, and evolving system behavior. Results on Raspberry Pi hardware may also not generalize to other edge platforms or deployment configurations.

\textbf{Construct Validity:}
RCA effectiveness is measured primarily using ranking-based metrics (AC@1, AC@5, AC@10, and AUC@10), which do not directly capture diagnostic interpretability or remediation effectiveness. Training time, memory, latency, and energy consumption provide practical proxies for edge efficiency but do not capture all deployment costs, such as communication, storage, and orchestration overhead. Comparisons between causal and non-causal approaches may also depend on dataset characteristics and training-data availability.

\section{Conclusion}

We presented SFlexRCA, a lightweight, scalable, and flexible root cause analysis framework for IIoT telemetry that avoids explicit dependency graph construction and message passing. Across BATADAL, SWAT, and WADI, SFlexRCA achieved competitive or superior RCA accuracy while maintaining a compact architecture. Its parameter requirements decreased on WADI despite having the largest number of monitored variables, demonstrating that SFlexRCA can adapt to different telemetry dimensionalities without parameter growth proportional to the number of variables. The efficiency evaluation further demonstrated low training memory and computational overhead and efficient inference on Raspberry Pi 3 and Raspberry Pi 5.
The results highlight an important trade-off in IIoT RCA. Non-causal forecasting-based approaches can provide efficient modeling but may require substantial normal-operation data, limiting their effectiveness under data-constrained regimes. Causal and dependency-aware approaches can capture relationships underlying fault propagation but may introduce substantial parameter and computational overhead through explicit dependency modeling, per-variable networks, or message passing. SFlexRCA addresses this trade-off by using orthogonal representations and shared lightweight linear modeling to capture cross-variable and temporal structure without explicitly constructing a dependency graph. This enables effective RCA under limited training data while avoiding parameter growth with increasing telemetry dimensionality.
The ablation and sensitivity analyses further demonstrate that the architecture, loss formulation, and temporal context can be adapted to different dataset characteristics. Representation comparisons and t-SNE visualization provide additional evidence that the learned orthogonal representations capture informative and separable structure for root cause localization. Overall, SFlexRCA provides a practical balance among RCA accuracy, training-data requirements, model complexity, and edge computational cost, enabling topology-free and resource-efficient root cause localization for IIoT systems.

\bibliographystyle{IEEEtran}
\bibliography{lipics-v2021-sample-article}



\end{document}